\documentclass{article}

\usepackage{arxiv}

\usepackage[utf8]{inputenc} % allow utf-8 input
\usepackage[T1]{fontenc}    % use 8-bit T1 fonts
\usepackage{url}            % simple URL typesetting
\usepackage{booktabs}       % professional-quality tables
\usepackage{amsfonts}       % blackboard math symbols
\usepackage{nicefrac}       % compact symbols for 1/2, etc.
\usepackage{microtype}      % microtypography
\usepackage{lipsum}
\usepackage{graphicx}
\graphicspath{ {./images/} }
\usepackage{float}

\title{Spatio-temporal Data Augmentation for Visual Surveillance}
\def\correspondingauthor{\footnote{Corresponding author : jeha@seoultech.ac.kr}}
\author{
 Jaeyeul Kim \\
  Graduate School of Automotive Engineering\\
  Seoul National University of Science and Technology\\
  Seoul, Korea \\
  \texttt{18512086@seoultech.ac.kr} \\
   \And
 Jongeun Ha \correspondingauthor{} \\
  Department of Mechanics and Automotive Engineering\\
  Seoul National University of Science and Technology\\
  Seoul, Korea \\
  \texttt{jeha@seoultech.ac.kr} \\
}

\begin{document}
\maketitle

\newcommand\blfootnote[1]{%
  \begingroup
  \renewcommand\thefootnote{}\footnote{#1}%
  \addtocounter{footnote}{-1}%
  \endgroup
}

\renewcommand{\thefootnote}{\asterisk{footnote}}
\blfootnote{*Corresponding author : jeha@seoultech.ac.kr}

\begin{abstract}
Visual surveillance aims to stably detect a foreground object using a continuous image acquired from a fixed camera. Recent deep learning methods based on supervised learning show superior performance compared to classical background subtraction algorithms. However, there is still a room for improvement in static foreground, dynamic background, hard shadow, illumination changes, camouflage, etc. In addition, most of the deep learning-based methods operates well on environments similar to training. If the testing environments are different from training ones, their performance degrades. As a result, additional training on those operating environments is required to ensure a good performance. Our previous work which uses spatio-temporal input data consisted of a number of past images, background images and current image showed promising results in different environments from training, although it uses a simple U-NET structure. In this paper, we propose a data augmentation technique suitable for visual surveillance for additional performance improvement using the same network used in our previous work. In deep learning, most data augmentation techniques deal with spatial-level data augmentation techniques for use in image classification and object detection. In this paper, we propose a new method of data augmentation in the spatio-temporal dimension suitable for our previous work. Two data augmentation methods of adjusting background model images and past images are proposed. Through this, it is shown that performance can be improved in difficult areas such as static foreground and ghost objects, compared to previous studies. Through quantitative and qualitative evaluation using SBI, LASIESTA, and our own dataset, we show that it gives superior performance compared to deep learning-based algorithms and background subtraction algorithms. In addition, it has a 30.2\% and 27.9\% reduction of false detection rate in the LASIESTA and SBI dataset, respectively, compared to our previous study. 
\end{abstract}

% keywords can be removed

\section{Introduction}
Static foreground, dynamic background, hard shadow, illumination changes, camouflage, bootstrap, bad weather, and occlusion, as shown in Fig. 1, are typical difficult problems to solve in visual surveillance. When we use change of brightness value to detect foreground objects, it causes difficulties in a stationary foreground object and a moving background portion because the former has little brightness change while the latter has large brightness change. In the case of hard shadow, it must be determined as a background object although there is a large brightness difference. Even in an environment where light changes rapidly, if only temporal information is relied on, it can cause serious false positives, as most BGS (Background Subtraction) algorithms show. In a camouflage environment in which the pixel values of the background object and the foreground object are similar, it is difficult to perform robust detection even if both spatial and temporal information are used. In the case of bootstrap where the foreground object exists at the beginning of the image sequence, a serious error may occur in initializing the background model. Weather factors such as snow and rain also act as a big challenge in visual surveillance. Simple algorithms may consider snow and rain as foreground objects. Recent methods using deep learning have improved performance to a certain degree for these difficult problems compared to traditional BGS-based methods, but research is still needed to improve performance. 

\begin{figure}[H]
  \centering
  \includegraphics[scale=0.25]{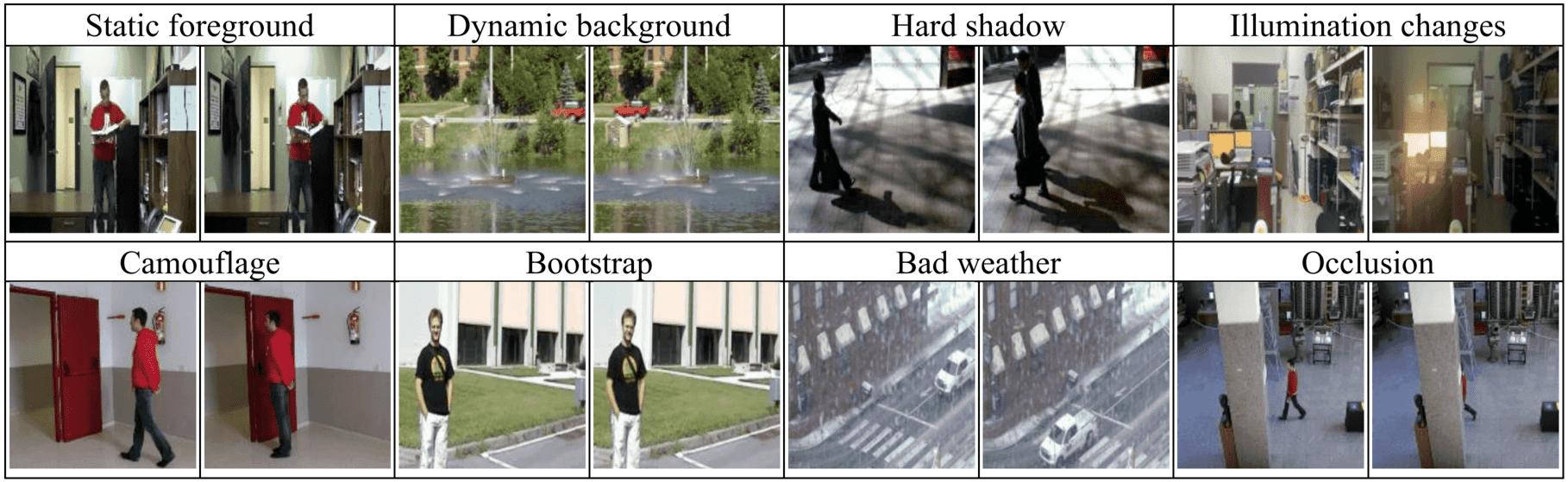} 
  \caption{Typical difficulties in visual surveillance.}
  \label{fig:fig1}
\end{figure}

Fig. 2 shows the problems that can occur when using only spatial information. A red rectangle shows a foreground object and a green rectangle shows a background object. If we only use spatial information without using temporal information, both cars may be detected as foreground objects. Since temporal information is lacking in a current image, it may be difficult to determine whether a vehicle on the image is stopped, parked, or moving. If a background model image has well reflected spatio-temporal information, we can correctly detect foreground and background objects. For this reason, it is essential to properly acquire spatio-temporal information from the background and past images in the visual surveillance. In our previous work [1], we proposed a visual surveillance algorithm which extracts spatio-temporal information by using the current image, the past images, and the background model image as the input of convolutional neural network as shown in Fig. 3. In most environments, it showed superior results compared to the latest deep learning-based algorithms and traditional BGS algorithms. However, unstable detection results were shown in static foreground and bootstrap environments. In this paper, we propose data augmentation methods to improve performance in static foreground and bootstrap environments.

\begin{figure}[H]
  \centering
  \includegraphics[scale=0.25]{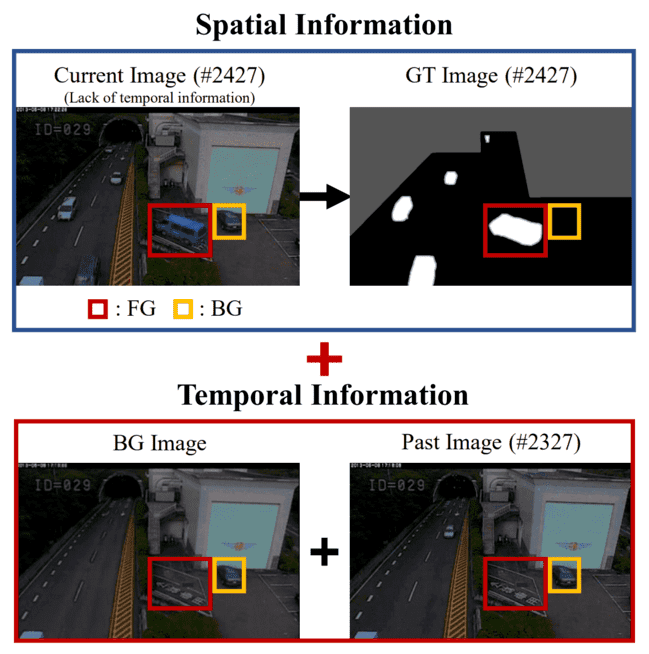}
  \caption{Limitations that can occur in visual surveillance using only spatial information.}
  \label{fig:fig2}
\end{figure}

\begin{figure}
  \centering
  \includegraphics[scale=.3]{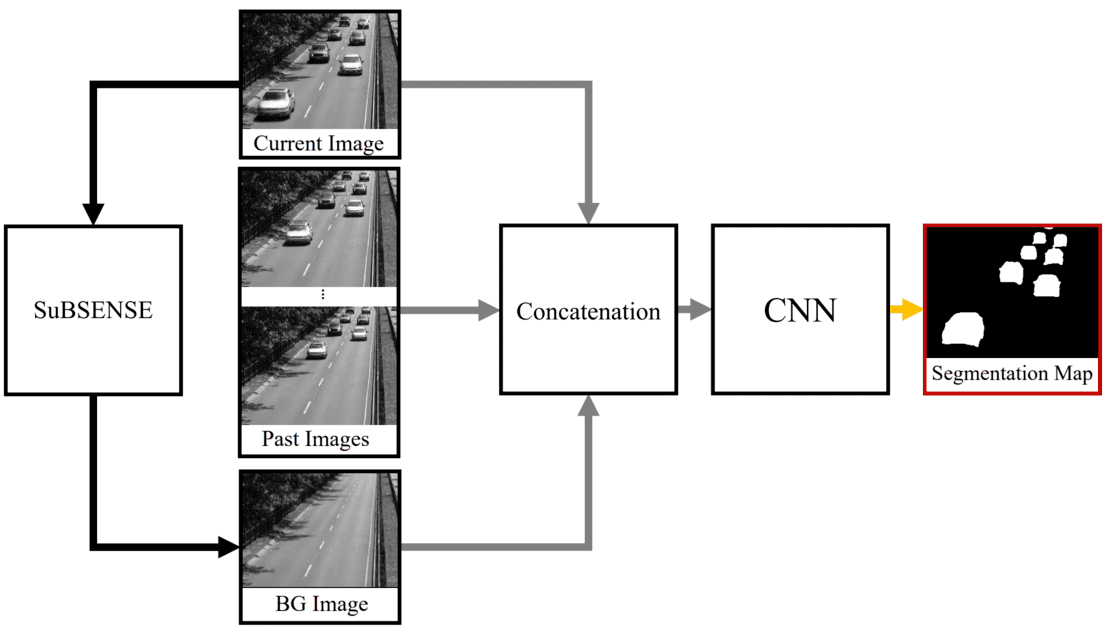}
  \caption{The network structure with multiple inputs used in our previous work [1].}
  \label{fig:fig3}
\end{figure}

\begin{figure}
  \centering
  \includegraphics[scale=.25]{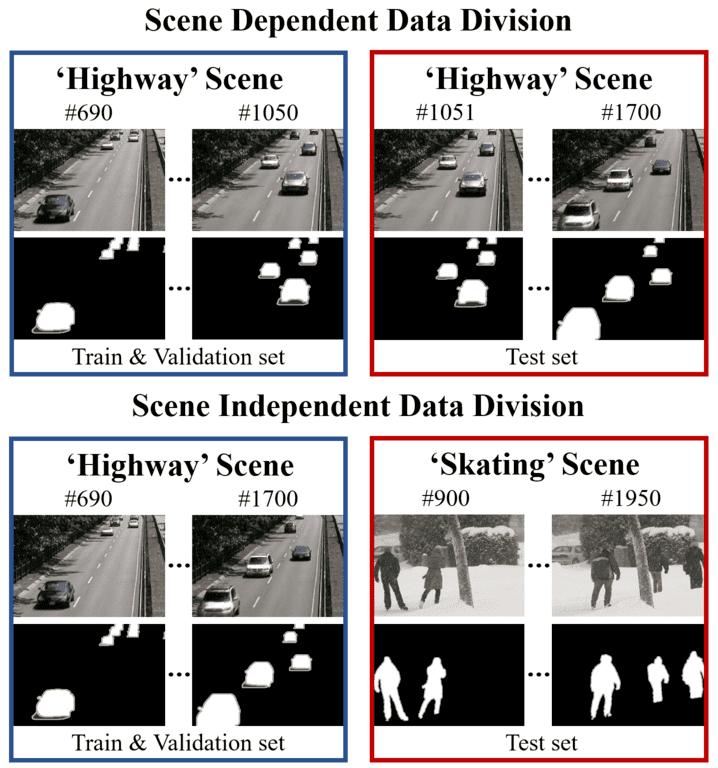}
  \caption{The illustration of scene dependent data division and scene independent data division.}
  \label{fig:fig4}
\end{figure}

The generalization ability is evaluated in two ways in visual surveillance. One is a method of dividing one image sequence into a train and test set, and the other is a method of separating an image sequence used for train and test. Mandal et al. [30] named these two methods as scene dependent data division (SDE) and scene independent data division (SIE), and we follow these notations in this paper. Fig. 4 shows a conceptual diagram of SDE and SIE setup. Most of the deep learning-based methods for visual surveillance train and evaluate under SDE setup. They show a good performance over the entire image sequence by using only a few ground-truth labels in a corresponding environment. However, even with a small number of ground-truth labels, generating them requires a lot of time and cost. Also, their performance is poor in environments different from the training environment. The generalization ability under SIE setup is more important than that of under SDE setup in visual surveillance if we consider their practical use in many diverse environments. In this paper, we focus on improving the generalization ability under SIE setup. We propose data augmentation methods suitable for spatio-temporal data which guarantee an improved performance under SIE setup. Data augmentation methods were devised through the investigation of difference in training results when using the CDnet2014 dataset and other datasets. We found that two items were main cause of the performance degradation through the analysis of the experimental results. First, the overall performance is degraded due to an error in the generated a background model image. Second, a person whose position is fixed in the image for more than a certain period of time causes a severe problem. In visual surveillance, people should always be detected as foreground objects regardless of whether they are moving.
Fig. 5 shows the different aspects of the CDnet [3] and LASIESTA [5] dataset. The CDnet dataset [3] which provides ground-truth labels for diverse environments is widely used in visual surveillance. It provides ground-truth labels only in the middle of each scene, therefore classical BGS algorithms can generate an appropriate level of background model image by using previous images at the location where ground-truth labels are given. However, in the SBI [4] and LASIESTA [5] datasets, there is no separate preparation section like CDnet dataset since ground-truth labels are given from the first image of the sequence. Therefore, classical BGS algorithms provide unstable background model images at the initial stage. Fig. 6 shows unstable background model images when foreground objects exists at the start position in an image sequence. In the CaVignal scene of the SBI dataset, the foreground object of the person stands still from 1 to 100 images, which makes it difficult to create correct background model images as shown in Fig. 6. Traditional BGS algorithms of SuBSENSE [2] and PAWCS [6] have difficulty in recovering correct background model images when foreground objects exist at fixed location. This is a problem that occurs frequently even in real situations of visual surveillance.

The main contributions of this paper are as follows:
\begin{itemize}
\item	We propose spatio-temporal data augmentation methods which are suitable for the network having multiple types of inputs in visual surveillance. Two methods of data augmentation are proposed. First, we propose a data augmentation method for background model image by intentionally adjusting the input images. Second, we propose a data augmentation method for improved detection of foreground objects of people by adjusting interval of images in an image sequence.
\item	The proposed data augmentation methods show clear improvement in two difficult cases of static foreground objects and bootstrap in visual surveillance. We present quantitative and qualitative experimental results using the CDnet, SBI, LASIESTA dataset and self-acquired dataset to show the improvement by the proposed data augmentation methods.
\item	We also show that we can obtain improved generalization power compared to previous algorithms. This is confirmed by experiments that we test trained model at environments different from training. For example, we test a model trained using CDnet dataset on the SBI dataset.
\end{itemize}

\begin{figure}[H]
  \centering
  \includegraphics[scale=.23]{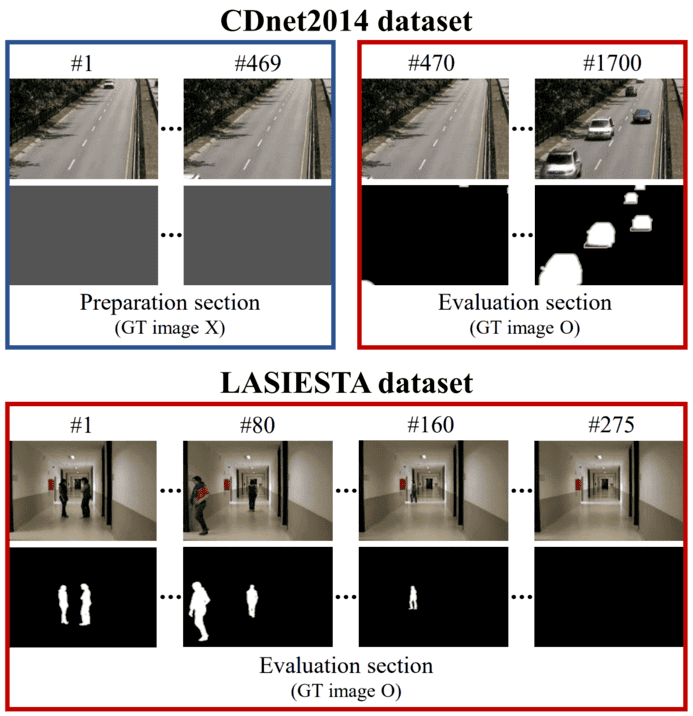}
  \caption{The difference of the configuration of CDnet and LASIESTA dataset.}
  \label{fig:fig5}
\end{figure}

\begin{figure}[H]
  \centering
  \includegraphics[scale=.23]{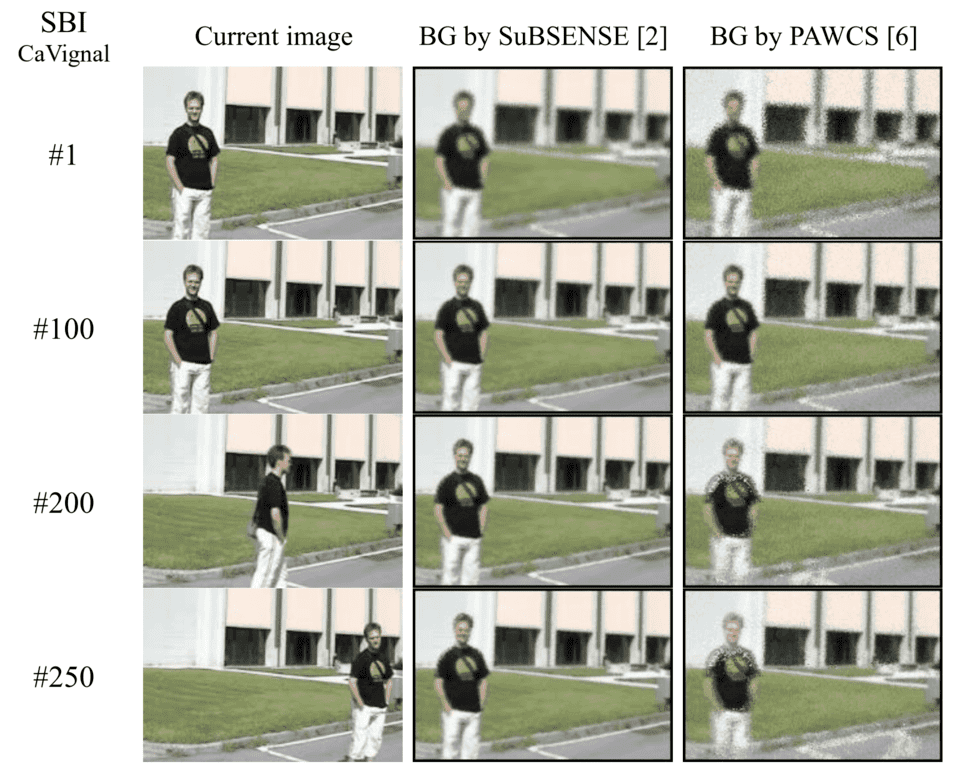}
  \caption{Examples of incorrectly generated background model images.}
  \label{fig:fig6}
\end{figure}

\section{Related works}
\label{sec:headings}
\subsection{Traditional approaches}

Stauffer and Grimson [24] proposed mixture of Gaussian (MOG) algorithm which represents the brightness value of each pixel as the combination of multiple Gaussian distributions. The number of the Gaussian mixture and each parameter of Gaussian distribution is determined by the expectation and maximization (EM) algorithm [40]. Special initialization is not required because it adapts their parameters as a sequence goes on. Pixels are considered as background when their brightness values belong to the Gaussian mixture model, otherwise, they are considered as foreground. Elgammal et al. [25] proposed a probabilistic non-parametric algorithm using kernel density estimation. Kim et al. [26] proposed an algorithm which uses a codebook. Codewords are established using intensity, color and temporal features, and they build up a codebook for later segmentation. Barnich et al. [14] proposed a background detection algorithm based on BGS. Foreground and background objects detection is performed for each pixel, and for a pixel determined as a background label, the pixel location or adjacent pixel is updated. In the foreground and background objects detection process, for each pixel, the 2-D Euclidean distance between the stored background model and the pixel value of the current input image is computed. Thereafter, the number of pixels of the background model image having a distance smaller than the threshold value is counted, and if the counted number is below the set threshold value, it is determined as a foreground. Haines et al. [11] proposed a new BGS algorithm based on non-parametric Bayesian density estimate. In addition, a new model update method called confidence capping that can degrade existing information in a principled way was proposed.

St-Charles et al. [2] proposed a BGS algorithm in which threshold values are automatically adjusted. Local distance thresholds and local update rates are adjusted according to the result of foreground and background objects detection. Local distance thresholds are set high in areas where dynamic backgrounds exist and local update rates are set high in areas where foreground objects are frequently detected. As a result, it is possible to respond to dynamic background motion and shows high performance among classical BGS algorithms. In the case of the background model generation part, it shows excellent performance in a general environment, but if there is a static foreground in the background model initialization step, there is a disadvantage that it takes a long time to recover the background model. Berjón et al. [10] proposed a nonparametric real-time and high quality foreground object detection method. The algorithm uses a KDE-based modeling strategy as its core, and a new Bayesian classifier is introduced. Ortego et al. [8] proposes a post-processing framework to improve the performance of BGS algorithms. They do motion-aware hierarchical segmentation using the segmentation result image of the BGS algorithms and the current image. Hierarchical foreground quality estimation and weighted foreground quality combination is done sequentially, and finally performing a foreground improvement is done. It was shown that the improved foreground map can be obtained. Yang et al. [12] proposes a spatiotemporally scalable matrix recovery (SSMR) method and a pyramid scheme for spatial scalability. Pyramid decomposition is used to separate foreground components at low resolution without down sampling the input image directly, and pyramid reconstruction is used to restore background components. Hossain et al. [9] proposed a non-smoothing color feature-based foreground detection method to effectively detect the foreground in a complex video environment. Thresholds are actively adjusted and in addition an adaptive post-processing method is proposed to cope with segmentation noises.

\subsection{Deep learning-based algorithms}

In the case of BGS algorithms, there is an advantage that a separate learning process is not required, but there are a number of parameters that need to be adjusted by the user. Parameters such as threshold and update rate are flexibly controlled through the feedback Scheme like SuBSENSE [2]. BGS algorithms are difficult to consider for the entire spatial image of the image beyond specific filter sizes such as 3x3, 5x5, and are vulnerable to problems such as static foreground, dynamic background, camouflage because they mainly rely on changes in pixel values. The latest deep learning-based methods are overcoming the limitations of these BGS algorithms and are showing improved performance.

Braham and Droogenbroeck [18] proposed a structure that decomposes the background image and the current image into a specific patch size and use them as the input of the CNN model. The background model image was extracted using 150 frames. After converting the RGB image to a gray image, a background image was generated by calculating the median value for each pixel of 150 images. Thereafter, an image patch is generated around each pixel of interest. This image patch generation process is performed identically to the background model image and the current image. After combining the two generated image patches, they are used as the input of the CNN which produces the foreground probability per pixel at the final stage. This method showed superior results compared to the BGS algorithms. However, the test was conducted in an SDE environment where 50\% of the front part of each scene in the dataset was used as train data and 50\% of the back part was used as test data. Zhao et al. [15] proposed a deep neural network consisting of two parts, a background reconstruction part and a foreground segmentation part. A model was proposed that receives the current image through the Background reconstruction process, generates a background image, and outputs the final foreground map by inputting the generated background image together with the current image into the FG segmentation part. This method shows superior performance compared to the BGS algorithms, but the test result was done in an SDE environment where half of each scene was divided into train and test. In addition, the foreground segmentation network does not accept past images, but only uses the background image and the current image generated by the background reconstruction network as inputs.

Wang et al. [27] proposed cascade CNN with two CNN models. The first CNN model receives the current image and obtains a foreground probability map, and the second CNN model stacks the probability map generated by the first CNN model and the current image and outputs a refined foreground probability map using as inputs. This method shows superior results compared to the BGS algorithms even if only 50 images are used for training in each scene, but the experiment in the SIE environment was not conducted. In addition, since only one current image is input, there is a limitation that the model cannot input information about temporal information. Zeng and Zhu [28] proposed a new multiscale fully convolutional network (MFCN) based on VGG-16 [22]. They replaced the complicated background modeling and updating process of classical algorithms with a simple network classification process. It shows excellent performance in the SDE environment, but it was not evaluated in the SIE environment. Babaee et al. [17] proposed a structure in which the background image and the current image are stacked depth-wise and they are used as the input to CNN model. They used a background pixel library by SuBSENSE foreground probability map to detect foreground objects. The proposed background model shows better performance than the background model of SuBSENSE [2]. This method shows excellent performance in the SDE environment, but shows relatively poor performance in the SIE environment. Lim et al. [31] proposed the multi-scale segmentation architectures FgSegNet-M and FgSegNet-S. Both structures receive a single image, and FgSegNet-M divides the image into three different scales, passes each through a CNN encoder, and decodes it using TCNN to obtain a final segmentation map. FgSegNet-S extracts multiple scales using the feature pooling module (FPM). Both methods show the best performance in the benchmark using the CDnet dataset. However, since FgSegNet is designed for the SDE environment, it is very vulnerable in the SIE environment. Lim et al. [16] proposed FgSegNet-v2, which uses pre-trained VGG16 [22] as an encoder part and generates a segmentation map using a decoder after passing through the modified FPM of the encoded features. FgSegNet-v2 currently ranks highest in the CDnet2014 dataset. However, like the existing FgSegNet-M and FgSegNet-S, since it is a structure that receives one image, it has a limitation of showing very poor performance in the SIE environment. Lin et al. [32] proposed a structure that concatenates the current image and the SuBSENSE background image and passes it through a model consisting of 20 convolutional layers and 3 deconvolutional layers to obtain a foreground map. The initial 7 convolution layers used VGG16 pre-trained weights. However, in the case of the model, since only the input image and the background image are input, it may be vulnerable to errors included in the background model. In our previous study [1], we showed that performance is improved when past images are together as input compared to the case of only the current image and background image are used as input.

Patil and Murala [7] proposed a compact end-to-end CNN model called motion saliency foreground network (MSFgNet). The MSFgNet is composed of motion saliency network (MSNet) and foreground network (FgNet). MSNet consists of a serial combination of a background estimation network and a saliency estimation network. This method does not require a separate background image and has the advantage of estimating the background image from the background estimation network. However, it has a disadvantage that the quality of the generated background image is lower than that of SuBSENSE [2]. FgNet is an encoder-decoder network, which receives the motion saliency output from MSNet as an input and outputs the final foreground map. Zeng et al. [19] proposed an RTSS framework that combines semantic segmentation and classical BGS algorithm. They used SuBSENSE as the BGS algorithm, and ICNET [13] and PSPNet [33] are used for the semantic segmentation. They showed FM score improvement of 6.87\% compared to SuBSENSE in the CDnet dataset. However, they did not show superior performance compared to other deep learning-based methods. Qui and Li proposed [21] a fully convolutional encoder-decoder spatial-temporal network (FCESNet). The structure consists of a feature encoder, a spatial-temporal information transmission module, and a feature decoder. It receives multiple images and outputs multiple segmentation maps, and uses the ConvLSTM layer-based spatial-temporal information transmission (STIT) module to obtain features of both spatial and temporal dimensions. Gomaa et al. [20] proposed an algorithm for counting moving vehicles. They first detects candidates of moving vehicles through the background difference method using CNN. Finally, they count vehicles using KLT tracker and K-means clustering. This method shows excellent performance regardless of the position of the vehicle in the image. Akilan et al. [23] proposed a model composed of 3D CNN-LSTM encoder and 3D CNN-LSTM decoder. Through this structure, both short spatio-temporal features and long spatio-temporal features can be considered. The evaluation was conducted in the SDE environment where the front 70\% of each scene on the CDnet was used for train and the latter 30\% was used for test.

Mandal et al. [30] proposed a structure showing excellent performance in both SIE and SDE environments. In this method, 50 consecutive images are input and a background image is estimated using a GRBE (Gradual Reduction Background Estimation) block, and the estimated background image and the current image are FSR (Foreground Saliency Reinforcement) and MScE (Multi-Schematic Encoder), MScD (Multi-Schematic Decoder), CFD (Compact Foreground Detection) blocks to create a final foreground map is proposed. The GRBE block uses a 3D convolution layer and has a structure in which pooling in the time domain is performed without pooling in the spatial dimension. In the FSR block, the difference between the estimated background image and the current image is processed, and the difference layer and the current image are respectively passed through the 3D conv. After that, a foreground map is obtained through an encoder-decoder structure. Mandal et al. [29] proposed a structure that exhibits excellent performance in SIE and SDE environments as shown in [30]. They estimate a background model image where 50 consecutive images are used as input to a depth reductionist background estimation (DRBE) block composed of multiple maximum multi-spatial receptive (MMSR). The background model image, median image and the current image are concatenated in the contrasting feature assimilation (CFA) block. After that, a structure for obtaining a final foreground map through an encoder-decoder structure was proposed. They shows superior results compared to other deep learning-based methods in SIE and SDE environments. However, the quality of the background image generated by DRBE is lower than that of SuBSENSE [2]. In our previous work [1], we proposed an algorithm for visual surveillance where a background model image by SuBSENSE [2] current image, and multiple past images are used as input together into the U-NET. We showed improved performance under the SIE environment using various open datasets. However, it has a fundamental problem that is difficult to cope with when the background image generation part is wrong.

\section{Proposed Data Augmentation Method for Visual Surveillance}

In deep learning, it is important to apply appropriate data augmentation to prevent model over-fitting and improve generalization ability. Data augmentation is widely used in image classification and object detection. There are geometric transform such as flip, rotation, shift, and zoom and photometric transform that adjusts the pixel value of an image using style transfer GAN such as CycleGAN [34, 35, 36, 37]. Since these methods treat one image not multiple consecutive images for data augmentation, they belong to data augmentation in the spatial domain. In our previous study [1], we proposed a method for visual surveillance having improved generalization power, which uses the current image, background model image, and past images as inputs to the network. In this paper, we propose a data augmentation method which could further improve our previous work. Data augmentation both in the spatial and in the temporal domain is required in visual surveillance. We propose spatio-temporal data augmentation methods by manipulating the background model image and past images.

\subsection{Data augmentation for background model image}

In our previous study [1], we used a background model image generated by SuBSENSE. It creates a suitable background model image in most cases. However, it cannot give a correct background model image in all cases. For example, in a bootstrap environment, generated background model image contains moving foreground objects as background. We manipulate the background model image on purpose to cope with error in the background model image. We propose a data augmentation method for the background model image in two directions. The first is when the BGS algorithm generates high quality background image. In this case, we propose a method to intentionally generate a wrong background model image. The other is when the BGS algorithm creates incorrect background images. In this case, we propose a method to generate a more correct background model image.

In our previous study [1], the background model image was sequentially generated from the first image of each scene using the SuBSENSE algorithm. BGS algorithms show an unstable appearance in the background model generation at the beginning. However, in the case of using CDnet2014 data, since SuBSENSE starts using hundreds or thousands of images before the time when label data is provided, a stable background image can be obtained at the time of training. For this reason, it was possible to perform training using the correct background images in the model training process. However, such a training data configuration increases the probability of rendering incorrect results if false background images are used as input when applied to an actual test environment after training. In order to cope with this situation, we additionally use background model images different from ones generated by SuBSENSE for training. When the SuBSENSE algorithm generates the correct background image, the order of the input images was changed to generate an incorrect background model image, and when the SuBSENSE algorithm generates the wrong background model image, the order of the input images was changed to generate the correct background image. These two types of background model images are used for training. This allows the data augmentation in visual surveillance. Most of the BGS algorithms receive the first image and use it to initialize the background model image. If dynamic foreground objects are included in the first image, they do not have a significant effect in the background model update process. But as shown in Fig. 7, when static foreground objects exist for a long time from the initial point, it takes a lot of time to restore the background model normally. In the proposed paper, this limitation of the BGS algorithm is used in the data augmentation technique in reverse. This situation was intentionally generated using the CDnet dataset, and we show that this type of data augmentation actually gives an improvement on visual surveillance by experiments.

\begin{figure}[H]
  \centering
  \includegraphics[scale=.25]{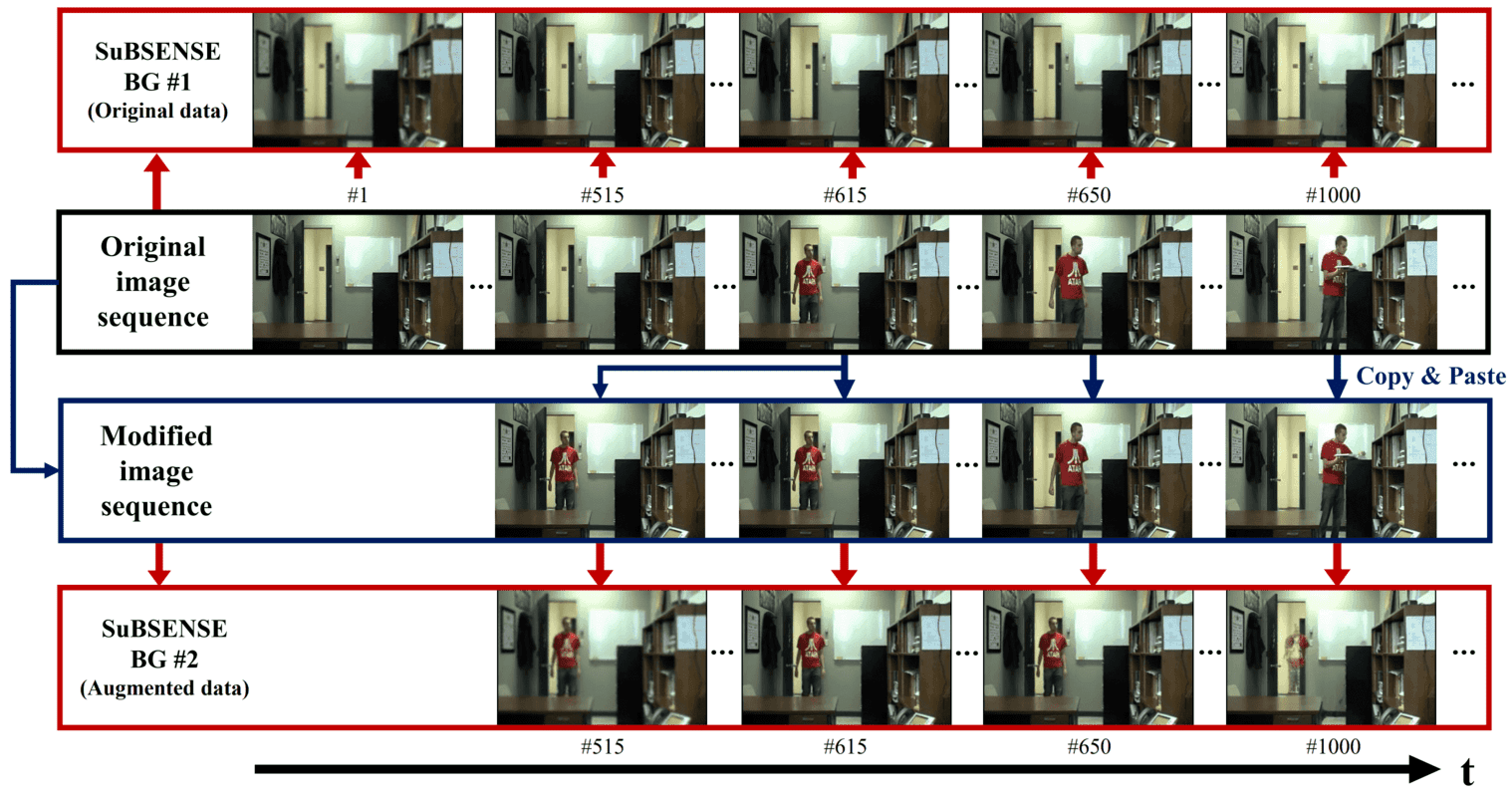}
  \caption{The procedure of background model image generation with intentional error.}
  \label{fig:fig7}
\end{figure}

\begin{figure}[H]
  \centering
  \includegraphics[scale=.3]{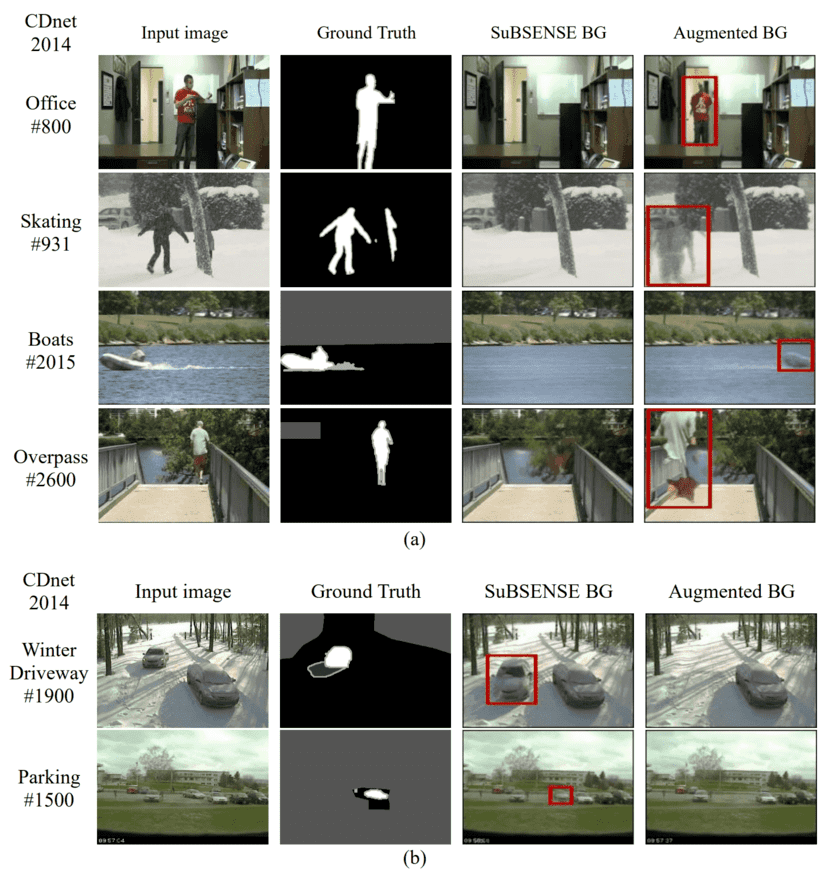}
  \caption{Result of data augmentation for a background model image (a) generation of wrong background model image (b) generation of correct background model image.}
  \label{fig:fig8}
\end{figure}

Augmenting a background model image is done in two ways. First, we generate wrong background model image on purpose. We modify the input image sequences used in the generation of the background model image. As shown in Fig. 7, we copied the image at the time the foreground object appeared and pasted it into the previous 100 indexes. In this way, we intentionally created the bootstrap environment and forced the BGS algorithm to generate an incorrect background image. Fig. 8(a) shows some examples of wrong background model image by this process. In Fig. 8(a), a red rectangle indicates the wrong area in the background model image. Second, we modify wrong background model image into correct one. Most of the BGS algorithms have a high probability of generating an incorrect background model image under the bootstrap environment or when there is a change in the background object. Fig. 8(b) shows that SuBSENSE generates incorrect background model images when parked vehicles are moved. Contrary to the previous method of generating background model images, accurate background model images were generated by pasting an image without foreground objects into the part from 100 indexes. The image that has no foreground objects was manually searched and selected within the entire image sequence. Throughout this process, we could obtain correct background model image. Fig. 8(b) shows some examples of corrected background model image when SuBSENSE gives a wrong background model image. In Fig. 8(b), a red rectangle indicates the wrong area in the background model image by SuBSENSE. These background model image modification methods have the disadvantage of manual processing, but they can effectively increase the number of training data.

\subsection{Data Augmentation by Adjusting Interval of Input Images}

In visual surveillance, foreground objects refer to the moving ones, but this definition does not apply to humans. Humans are considered as foreground objects regardless of movement, and this contradiction makes stable detection of them difficult in visual surveillance. In the case of CDnet, SBI, and LASIESTA datasets, humans are classified as foreground objects regardless of whether they are moving or not while moving vehicles are classified as foreground objects and parked vehicles are classified as background.

Detecting a person who has stopped for a long time as a foreground object is a difficult problem to solve in visual surveillance. In our previous study [1], the current image, background model image and previous images are used as the input of a network. In this paper, we propose a data augmentation method by adjusting the frame interval of past images. When data augmentation is not used, previous images are automatically selected according to a predetermined frame interval. In case of data augmentation, the frame intervals of previous images are all set to 0 as shown in Fig. 9. This corresponds to replacing all the past images with the same images as the current image. In this way, augmentation of the overall spatio-temporal information is possible by simply changing the composition of the input images, without directly changing the spatial dimension information of each image. In this paper, 27 scenes from the CDnet 2014 dataset in Table 1 are used for training. In Table 1, 11 scenes among 27 scenes are marked in bold correspond to scenes containing human objects not vehicles as foreground objects. Interval augmentation is applied only to these 11 scenes. In our previous work [1], 4 previous images selected at intervals of 25 frames in the range of 100 frames were used as input of a network. In the case of humans, even when there is no movement, it should be recognized as a foreground objects. Therefore, new data can be generated by changing the existing 25 frame interval to 0 frame interval. Since this method has the same ground truth image and simply changes the frame interval, there is no additional cost. In this way, we can obtain an additional 2,200 training sets from 11 scenes. Finally, the training data can be amplified by more than two times through two augmentation methods. From 5,400 images of 27 scenes, 7,600 learning sets can be obtained through interval augmentation. In addition, training data is amplified by 2 times using background model image augmentation. Thus, a total of 15,200 training sets can be used for training. Both data augmentation methods have the disadvantage of requiring manual intervention, but there is an advantage of securing detection performance improvement through this.

\begin{figure}[H]
  \centering
  \includegraphics[scale=.35]{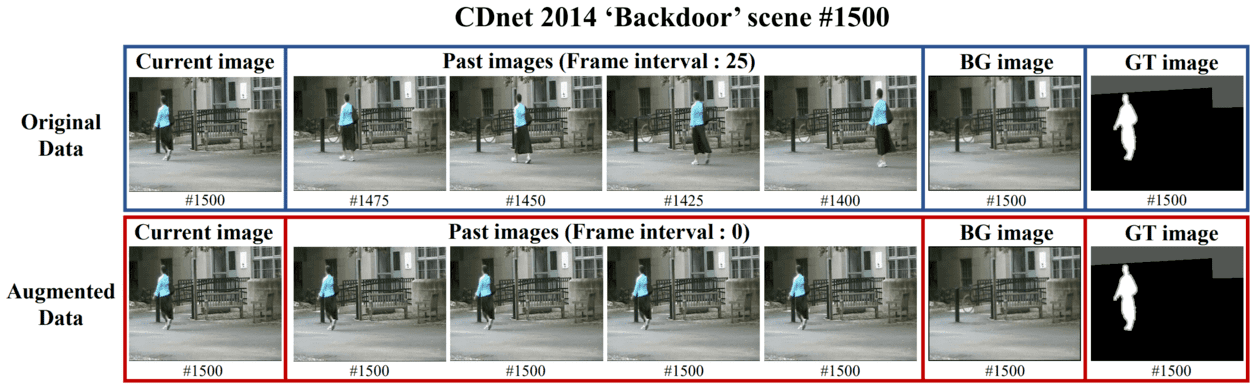}
  \caption{Data augmentation by adjusting the interval of previous images.}
  \label{fig:fig9}
\end{figure}

\begin{table}[]
\caption{Configuration of CDnet dataset (bold includes only foreground objects of people).}
\centering
\begin{tabular}{|c|c|}
\hline
Categories                 & Scenes                                                                      \\ \hline
Baseline                   & \textbf{Office}, \textbf{PETS2006}, Highway, \textbf{Pedestrians}                             \\ \hline
Bad Weather                & Wet snow, Blizzard, \textbf{Skating}, Snowfall                                       \\ \hline
Dynamic Background         & Boats, Canoe, Fountain1,   Fountain2, Fall, \textbf{Overpass}                        \\ \hline
Intermittent Object Motion & Streetlight, Parking, Winter   drive way, \textbf{Sofa}                              \\ \hline
Low Framerate              & Tunnel exit\_0\_35fps, Tram crossroad\_1fps, Turnpike\_0\_5fps              \\ \hline
Shadow                     & \textbf{Copy machine}, Bungalows, \textbf{Bus station}, \textbf{Back   door}, \textbf{Cubicle}, \textbf{People in shade} \\ \hline
\end{tabular}
\end{table}

\section{Experimental results}
\label{sec:others}

In experiments, we use the same network from our previous work [1] to show improved performance by proposed data augmentation methods. Fig. 10 shows the U-Net used in experiments where only the input image size has been changed from 320(W)X240(H)X6(C) to 224(W)X224(H)X6(C). The image size was changed to the same input size as VGG-16 in order to use VGG-16 pre-trained weights in later studies. The image was preprocessed by subtraction by 127.5 then dividing by 255. The number of parameters of the model is 31,098,049 and the number of trainable parameters is 31,084,225. Computation is done in 15ms with a PC having NVIDIA RTX 2080Ti, so it is sufficient to use for visual surveillance in a real-time. During training, 80\% of the total images was used for train and remaining 20\% is used for validation set. Binary cross entropy was used for loss and He initialization [38] was used for kernel initialization. Adam optimizer [39] was used as an optimizer. The initial learning rate was set to 0.001, and if the validation loss did not decrease more than 5 times, the learning rate was reduced by half. The training was stopped when the validation loss did not decrease more than 10 times. The last output layer is obtained by sigmoid, therefore it has an output between 0 and 1. For quantitative evaluation of foreground and background detection, values above 0.5 were set to 1 and values below 0.5 were set to 0.

Quantitative and qualitative analysis are done using the CDnet, SBI, LASIESTA datasets and one acquired ourselves. Most of deep learning-based visual surveillance algorithms are evaluated in SDE environment. Providing good results in the SDE environment can also be regarded as a kind of generalization ability. However, when they are applied to an environment different from the training environment, it provides even inferior results than the conventional BGS methods. In order to obtain better results than the traditional BGS algorithms, it is necessary to manually generate training data from the new environment. This paper aims to obtain superior results than the existing BGS algorithms when applied to a new environment without re-training. Training was done in two ways to show the generalization ability of proposed algorithm in SIE environment. First, we train the model using the CDnet dataset then we evaluate it on the SBI, LASIESTA, and self-obtained datasets. Second, we make SIE environment by dividing the SBI and LASIESTA dataset internally. For the quantitative evaluation, 27 scenes from the CDnet2014 dataset described in Table 1 are used for training. We use the same 200 images for each scene for training which are used in used in FgSegNet-v2 [16] except the sofa scene where training images are manually selected. We present comparison results depending on whether the proposed data augmentation method is used. Quantitative evaluation was done using the SBI and LASIESTA DATASET, and only qualitative evaluation was done for our own dataset. In visual surveillance, it is appropriate to judge moving bushes as background objects, therefore the Snellen and Foliage scenes in the SBI dataset that classify moving leaves as foreground objects are excluded from the evaluation. Also, the Toscana scene was excluded from the evaluation because it consisted of only 6 images that were not consecutive. In the LASIESTA Dataset, the evaluation was done using 20 scenes obtained with fixed cameras. We evaluate the performance of visual surveillance algorithms using F-measure (FM), percentage of wrong classification (PWC), recall, precision, false positive rate (FPR), false negative rate (FNR), and Specificity (Sp) which are defined as follows.

\begin{center}
$ Precision = \frac{TP}{TP+FP} $ \\
$ Recall = \frac{TP}{TP+FN} $ \\
$ FM = \frac{2XPrecisionXRecall}{Precision+Recall} $ \\
$ PWC = \frac{FP+FN}{TP+TN+FP+FN} $ \\
$ FPR = \frac{FP}{TN+FP} $ \\
$ FNR = \frac{FN}{TP+FN} $ \\
$ Sp = \frac{TN}{TN+FP} $
\end{center}

\begin{figure}
  \centering
  \includegraphics[scale=.15]{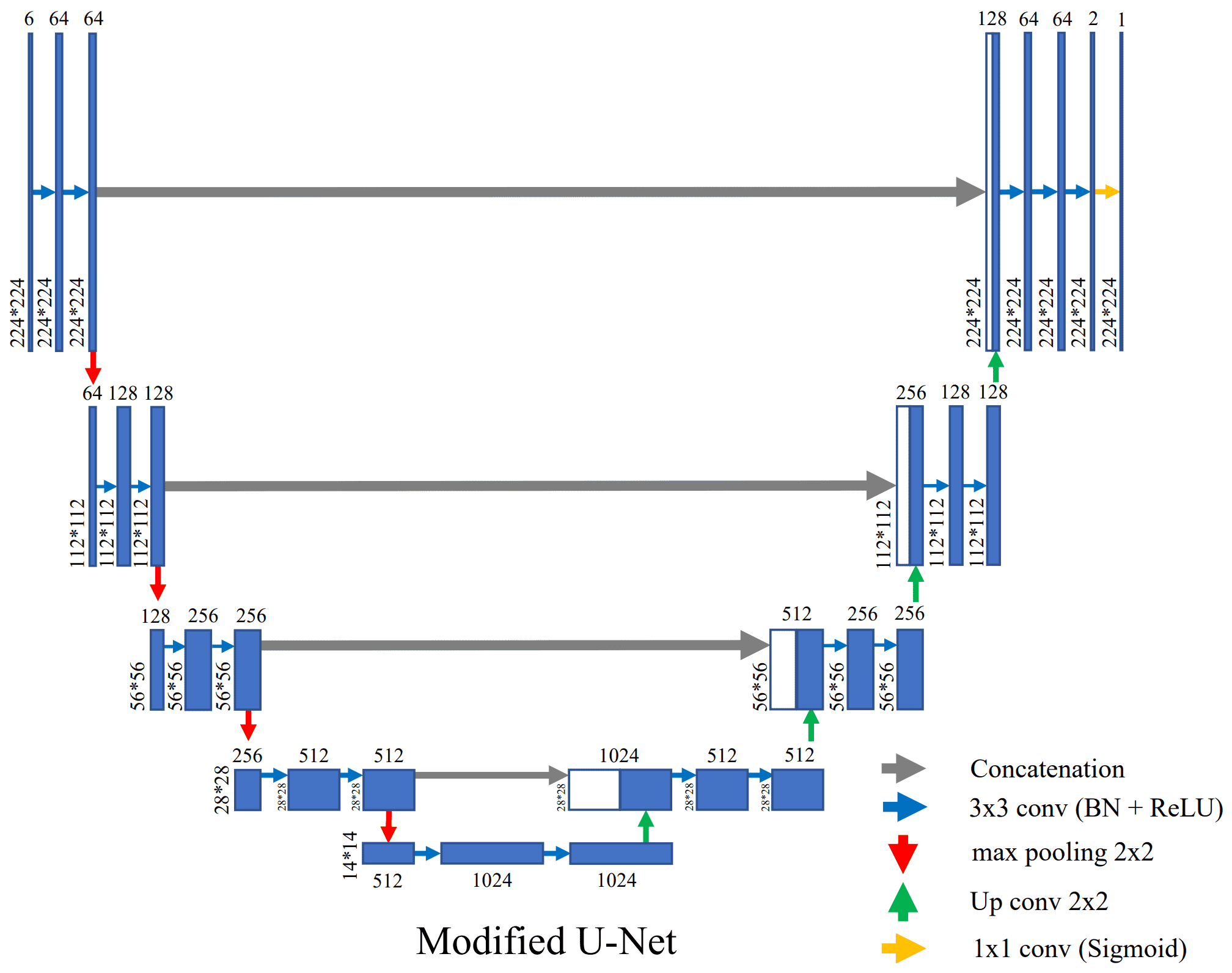}
  \caption{The network structure used in the experiments.}
  \label{fig:fig10}
\end{figure}

\subsection{Evaluation using training results with the CDnet dataset}

After training using the CDnet dataset, the trained model was tested on the SBI and LASIESTA datasets without re-training to evaluate the generalization ability in the SIE environment. First, evaluation was performed according to the use of the proposed data augmentation methods. Next, quantitative and qualitative evaluation was done by comparing to the latest deep learning-based algorithms and BGS algorithms. Only qualitative evaluation was done for the self-acquisition dataset.

\subsubsection{Evaluation on the SBI dataset}

First, quantitative and qualitative evaluation is done using the SBI dataset. Table 2 shows the results according to whether data augmentation methods are applied. The case where the augmentation method was not used is our previous work [1]. We can observe noticeable improvement only using background data augmentation compared to our previous work. Only applying interval data augmentation to our previous work decreases performance. Applying both background and interval data augmentation gives the best result with 7.0\% improvement in FM score and 27.9\% decrease in PWC score compared to our previous work.

Fig. 11 shows comparison results of foreground segmentation maps by various algorithms. From People\&Foliage \#22 and \#260 in Fig. 11, we can notice that using both background and interval data augmentation better detects people and recognizes moving leaves as background than using only background data augmentation. In the People\&Foliage scene of the SBI dataset, moving leaves are classified as foreground objects contrary to the usual convention of visual surveillance that they should be classified as background objects. This is controversial, therefore we clearly show the performance with and without People\&Foliage scene. Table 3 shows quantitative comparison results to other algorithms via FM score. Proposed algorithm shows better performance than traditional BGS algorithms even without using the SBI dataset for training. 3DCD [30] is a deep learning-based algorithm which targets its use in both SIE and SDE environments. We train 3DCD [30] using the same CDnet dataset, then test is done using the SBI dataset. Proposed algorithm gives 15.96\% improvement compared to 3DCD [30]. FgSegNet-v2 [16] shows very poor results in the SIE environment because it is a model designed for the SDE environment. Under the SDE environment, FgSegNet-v2 was trained using 5,400 images from 27 scenes in the CDnet dataset.  When evaluation was done using all images of 27 scenes in the CDnet dataset used for training, the average FM and PWC was 0.9838 and 0.0966, respectively, which amount to a good performance. Evaluation on the SBI dataset that was not used for training, it gives the average FM of 0.61, which was inferior to the BGS algorithms. From this, it can be seen that even the best results in the SDE environment are not suitable in the SIE environment. Table 4 shows detailed quantitative evaluation results by the proposed method.

\begin{table}[H]
\caption{FM and PWC values using the SBI dataset according to data augmentation cases.}
\centering
\small{
\begin{tabular}{ccccccccc}
\hline
                      & \multicolumn{2}{c}{\begin{tabular}[c]{@{}c@{}}No Aug {[}1{]}\\ (Total 5,400 images)\end{tabular}} & \multicolumn{2}{c}{\begin{tabular}[c]{@{}c@{}}BG Aug\\ (Total 10,800 images)\end{tabular}} & \multicolumn{2}{c}{\begin{tabular}[c]{@{}c@{}}Interval Aug\\ (Total 7,600 images)\end{tabular}} & \multicolumn{2}{c}{\begin{tabular}[c]{@{}c@{}}BG + Interval Aug\\ (Total 15,200 images)\end{tabular}} \\ \hline
                & FM                                              & PWC                                             & FM                                          & PWC                                          & FM                                             & PWC                                            & FM                                                & PWC                                               \\ \hline
Board                 & 0.9199                                          & 4.6148                                          & \textbf{0.9534}                             & \textbf{2.6783}                              & 0.9069                                         & 5.6739                                         & 0.9235                                            & 4.3153                                            \\ \hline
CIVIAR1               & 0.9477                                          & 0.4021                                          & 0.9513                                      & 0.3767                                       & 0.9410                                         & 0.4552                                         & \textbf{0.9521}                                   & \textbf{0.3673}                                   \\ \hline
CIVIAR2               & \textbf{0.9285}                                 & \textbf{0.0559}                                 & 0.9172                                      & 0.0662                                       & 0.8988                                         & 0.0821                                         & 0.9256                                            & 0.0586                                            \\ \hline
CaVignal              & 0.4371                                          & 10.9914                                         & 0.6763                                      & 5.8335                                       & 0.4228                                         & 10.8120                                        & \textbf{0.8145}                                   & \textbf{3.7146}                                   \\ \hline
Candela               & 0.7461                                          & 1.6913                                          & 0.7304                                      & 1.7155                                       & 0.7702                                         & 1.5459                                         & \textbf{0.8111}                                   & \textbf{1.2652}                                   \\ \hline
Hall \& Monitor       & 0.9567                                          & 0.2025                                          & 0.9704                                      & 0.1403                                       & 0.9618                                         & 0.1797                                         & \textbf{0.9713}                                   & \textbf{0.1361}                                   \\ \hline
Highway1              & 0.9123                                          & 1.7373                                          & 0.9341                                      & 1.2950                                       & 0.9228                                         & 1.5530                                         & \textbf{0.9439}                                   & \textbf{1.0731}                                   \\ \hline
Highway2              & 0.9711                                          & 0.1733                                          & 0.9717                                      & 0.1699                                       & 0.9730                                         & 0.1619                                         & \textbf{0.9741}                                   & \textbf{0.1529}                                   \\ \hline
Human Body2           & 0.9283                                          & 1.4564                                          & \textbf{0.9593}                             & \textbf{0.8139}                              & 0.9227                                         & 1.5884                                         & 0.9525                                            & 0.9608                                            \\ \hline
IBM TEST2             & 0.9627                                          & 0.3527                                          & 0.9779                                      & 0.2028                                       & 0.9685                                         & 0.2912                                         & \textbf{0.9821}                                   & \textbf{0.1635}                                   \\ \hline
People \& Foliage     & 0.5442                                          & 34.9605                                         & \textbf{0.6514}                             & \textbf{28.4632}                             & 0.5748                                         & 33.5349                                        & 0.6506                                            & 28.6471                                           \\ \hline
Average               & 0.8413                                          & 5.1489                                          & 0.8812                                      & 3.7959                                       & 0.8421                                         & 5.0798                                         & \textbf{0.9001}                                   & \textbf{3.7141}                                   \\ \hline
Average (Except P\&F) & 0.8710                                          & 2.1678                                          & 0.9042                                      & 1.3292                                       & 0.8689                                         & 2.2343                                         & \textbf{0.9251}                                   & \textbf{1.2208}                                   \\ \hline
\end{tabular}
}
\end{table}

% \begin{table}
%  \caption{FM and PWC values using the SBI dataset according to data augmentation cases.}
%   \centering
%   \begin{tabular}{lll}
%     \toprule
%     \multicolumn{2}{c}{Part}                   \\
%     \cmidrule(r){1-2}
%     Name     & Description     & Size ($\mu$m) \\
%     \midrule
%     Dendrite & Input terminal  & $\sim$100     \\
%     Axon     & Output terminal & $\sim$10      \\
%     Soma     & Cell body       & up to $10^6$  \\
%     \bottomrule
%   \end{tabular}
%   \label{tab:table}
% \end{table}

% \caption{The comparison result of FM score on the SBI dataset.}
% \centering

\begin{table}[H]
\caption{The comparison result of FM score on the SBI dataset.}
\centering
\small{
\begin{tabular}{cccccccc}
\hline
\begin{tabular}[c]{@{}c@{}}SBI\\ DATASET\end{tabular}             & \begin{tabular}[c]{@{}c@{}}Proposed\\ (SIE)\end{tabular} & \begin{tabular}[c]{@{}c@{}}Kim\&Ha\\ {[}1{]} (SIE)\end{tabular} & \begin{tabular}[c]{@{}c@{}}Yang et al.\\ {[}12{]}\end{tabular} & \begin{tabular}[c]{@{}c@{}}3DCD\\ {[}30{]} (SIE)\end{tabular} & \begin{tabular}[c]{@{}c@{}}PAWCS\\ {[}6{]}\end{tabular} & \begin{tabular}[c]{@{}c@{}}SuBSENSE\\ {[}2{]}\end{tabular} & \begin{tabular}[c]{@{}c@{}}FgSegNet-v2\\ {[}16{]} (SIE)\end{tabular} \\ \hline
Board                                                             & \textbf{0.9235}                                          & 0.9199                                                          & 0.91                                                           & 0.9094                                                        & 0.7798                                                  & 0.5777                                                     & 0.7792                                                               \\ \hline
CIVIAR1                                                           & \textbf{0.9521}                                          & 0.9477                                                          & \textbf{0.95}                                                           & 0.9428                                                        & 0.8589                                                  & 0.9144                                                     & 0.6346                                                               \\ \hline
CIVIAR2                                                           & 0.9256                                                   & \textbf{0.9285}                                                 & 0.84                                                           & 0.8147                                                        & 0.6772                                                  & 0.8714                                                     & 0.0577                                                               \\ \hline
CaVignal                                                          & 0.8145                                          & 0.4371                                                          & 0.83                                                           & 0.6186                                                        & 0.3697                                                  & 0.3980                                                     & \textbf{0.8576}                                                              \\ \hline
Candela                                                           & 0.8111                                                   & 0.7461                                                          & \textbf{0.93}                                                           & 0.5600                                                        & 0.8725                                         & 0.5356                                                     & 0.3370                                                               \\ \hline
Hall \& Monitor                                                   & \textbf{0.9713}                                          & 0.9567                                                          & 0.83                                                           & 0.7283                                                        & 0.7411                                                  & 0.7758                                                     & 0.7060                                                               \\ \hline
Highway1                                                          & \textbf{0.9439}                                          & 0.9123                                                          & 0.72                                                           & 0.7908                                                        & 0.7015                                                  & 0.5523                                                     & 0.8780                                                               \\ \hline
Highway2                                                          & \textbf{0.9741}                                          & 0.9711                                                          & 0.95                                                           & 0.6669                                                        & 0.9031                                                  & 0.8937                                                     & 0.7409                                                               \\ \hline
HumanBody2                                                        & \textbf{0.9525}                                          & 0.9283                                                          & 0.88                                                           & 0.8058                                                        & 0.7013                                                  & 0.8346                                                     & 0.5825                                                               \\ \hline
IBM TEST2                                                         & \textbf{0.9821}                                          & 0.9627                                                          & 0.89                                                           & 0.8725                                                        & 0.9386                                                  & 0.9390                                                     & 0.5717                                                               \\ \hline
People \& Foliage                                                 & 0.6506                                                   & 0.5442                                                          & -                                                              & \textbf{0.8290}                                               & 0.3162                                                  & 0.2660                                                     & 0.5644                                                               \\ \hline
Average                                                           & \textbf{0.9001}                                          & 0.8413                                                          & -                                                              & 0.7762                                                        & 0.7145                                                  & 0.6871                                                     & 0.6100                                                               \\ \hline
\begin{tabular}[c]{@{}c@{}}Average\\ (No P\&F scene)\end{tabular} & \textbf{0.9251}                                          & 0.8710                                                          & 0.87                                                           & 0.7710                                                        & 0.7544                                                  & 0.7293                                                     & 0.6145                                                               \\ \hline
\end{tabular}
}
\end{table}

\begin{table}[H]
\caption{Results by proposed algorithm on the SBI dataset.}
\centering
\small{
\begin{tabular}{cccccccc}
\hline
Scene                                                                 & FM     & PWC     & Recall & Precision & FPR    & FNR    & Sp     \\ \hline
Board                                                                 & 0.9235 & 4.3153  & 0.9698 & 0.8815    & 0.0479 & 0.0302 & 0.9521 \\ \hline
CIVIAR1                                                               & 0.9521 & 0.3673  & 0.9288 & 0.9769    & 0.0009 & 0.0712 & 0.9991 \\ \hline
CIVIAR2                                                               & 0.9256 & 0.0586  & 0.9153 & 0.9363    & 0.0002 & 0.0847 & 0.9998 \\ \hline
CaVignal                                                              & 0.8145 & 3.7146  & 0.9467 & 0.7151    & 0.0355 & 0.0533 & 0.9645 \\ \hline
Candela                                                               & 0.8111 & 1.2652  & 0.9884 & 0.6921    & 0.0127 & 0.0116 & 0.9874 \\ \hline
Hall \& Monitor                                                       & 0.9713 & 0.1361  & 0.9841 & 0.9588    & 0.0010 & 0.0159 & 0.9990 \\ \hline
Highway1                                                              & 0.9439 & 1.0731  & 0.9757 & 0.9142    & 0.0093 & 0.0243 & 0.9907 \\ \hline
Highway2                                                              & 0.9741 & 0.1529  & 0.9838 & 0.9647    & 0.0011 & 0.0162 & 0.9989 \\ \hline
Human Body2                                                           & 0.9525 & 0.9608  & 0.9571 & 0.9480    & 0.0059 & 0.0429 & 0.9941 \\ \hline
IBM TEST2                                                             & 0.9821 & 0.1635  & 0.9862 & 0.9779    & 0.0011 & 0.0138 & 0.9989 \\ \hline
\begin{tabular}[c]{@{}c@{}}People \&\\ Foliage\end{tabular}           & 0.6506 & 28.6471 & 0.9004 & 0.5095    & 0.3650 & 0.0996 & 0.6350 \\ \hline
Average                                                               & 0.9001 & 3.7141  & 0.9578 & 0.8614    & 0.0437 & 0.0422 & 0.9563 \\ \hline
\begin{tabular}[c]{@{}c@{}}Average\\(Except P\&F)\end{tabular} & 0.9251 & 1.2208  & 0.9636 & 0.8966    & 0.0116 & 0.0364 & 0.9884 \\ \hline
\end{tabular}
}
\end{table}

\begin{figure}[H]
  \centering
  \includegraphics[scale=.45]{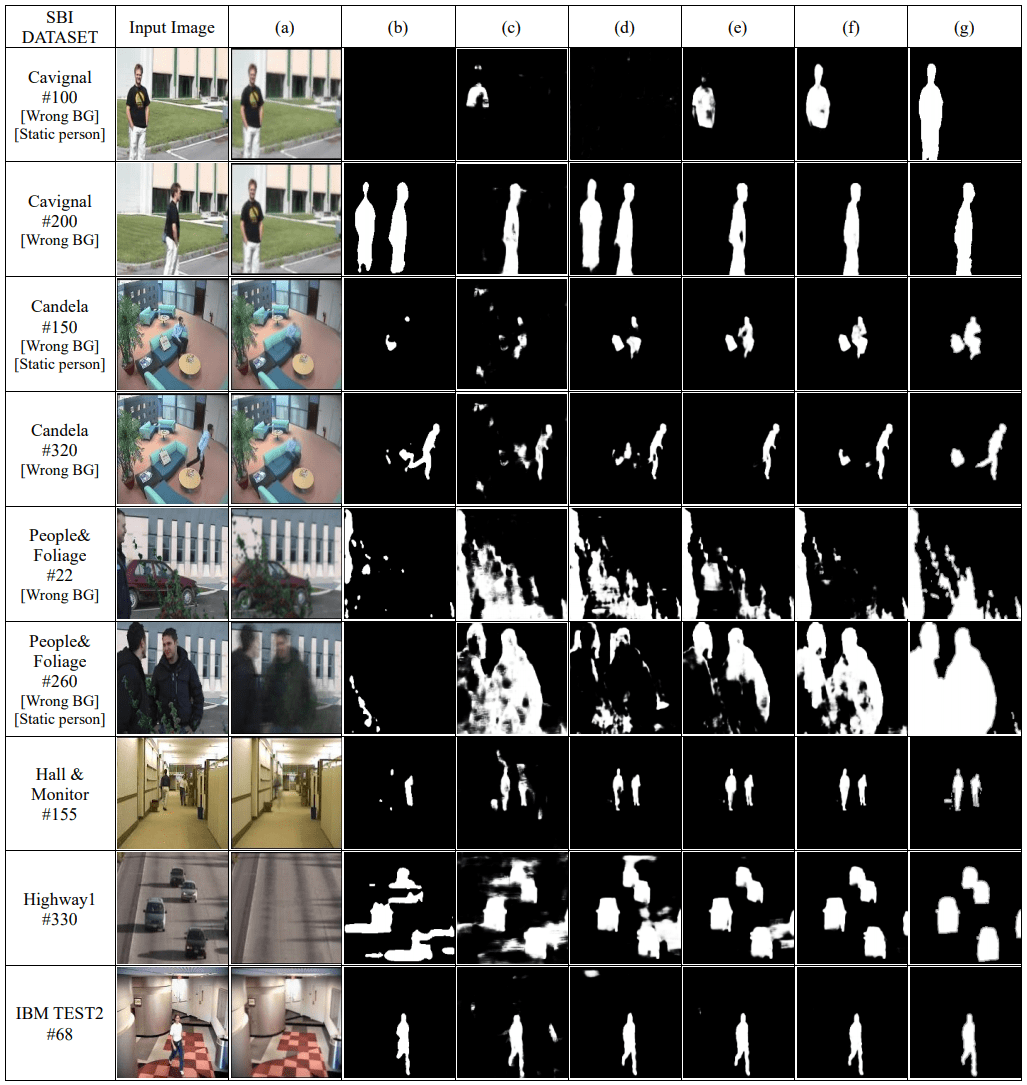}
  \caption{The comparison to other algorithms using SBI dataset (a) background model image by SuBSENSE [2] (b) foreground map by SuBSENSE [2] (c) foreground map by 3DCD [30] (d) foreground map by our previous work [1] (e) foreground map by proposed algorithm using only background data augmentation (f) foreground map by proposed algorithm using both background and interval data augmentation (g) ground truth foreground map.}
  \label{fig:fig11}
\end{figure}

\subsubsection{Evaluation on the LASIESTA dataset}

In the LASIESTA dataset, evaluation was done by dividing into 4 cases according to whether or not augmentation was applied like experiments on the SBI dataset. Table 5 shows the quantitative evaluation results according to the application of the proposed data augmentation methods. We can obtain FM score of 0.93 which is a good value without using data augmentation for the LASIESTA dataset. We cannot observe distinct improvement like in the SBI dataset because the LASIESTA dataset contains few ghost objects compared to the SBI dataset. When we use both background and interval data augmentation, PWC score decreases from 0.2966 to 0.2070 which amount to 30.21\% decrease. In the bootstrap category of I-BS where the foreground object exists from the first image, FM score improves from 0.7892 to 0.8758 which amount to by 10.97\% improvement and PWC decreased from 0.4273 to 0.2250 which amount to 47.34\% decrease. The proposed data augmentation methods can cope with problems caused by incorrect background model images. Fig. 12 shows some representative results for each scene in the LASIESTA dataset by various algorithms. The proposed algorithm gives the best result for detecting a stationary human as can be seen in Fig. 12. Table 6 shows quantitative comparison results to other algorithms. Table 7 shows various quantitative statistics by the proposed algorithm using the LASIESTA dataset.

\begin{table}[H]
\caption{FM and PWC values using LASIESTA dataset according to data augmentation cases.}
\centering
\small{
\begin{tabular}{ccccccccc}
\hline
        & \multicolumn{2}{c}{\begin{tabular}[c]{@{}c@{}}No Aug {[}1{]}\\ (Total 5,400 images)\end{tabular}} & \multicolumn{2}{c}{\begin{tabular}[c]{@{}c@{}}BG Aug\\ (Total 10,800 images)\end{tabular}} & \multicolumn{2}{c}{\begin{tabular}[c]{@{}c@{}}Interval Aug\\ (Total 7,600 images)\end{tabular}} & \multicolumn{2}{c}{\begin{tabular}[c]{@{}c@{}}BG + Interval Aug\\ (Total 15,200 images)\end{tabular}} \\ \hline
        & FM                                              & PWC                                             & FM                                           & PWC                                         & FM                                               & PWC                                             & FM                                                & PWC                                               \\ \hline
I\_SI   & 0.9617                                          & 0.2229                                          & 0.9631                                       & 0.2206                                      & \textbf{0.9664}                                  & \textbf{0.1946}                                 & 0.9301                                            & 0.3263                                            \\ \hline
I\_CA   & 0.9145                                          & 1.0482                                          & 0.8935                                       & 1.2831                                      & \textbf{0.9660}                                  & \textbf{0.4047}                                 & 0.9589                                            & 0.5089                                            \\ \hline
I\_OC   & 0.9765                                          & 0.1235                                          & 0.9737                                       & 0.1279                                      & \textbf{0.9774}                                  & \textbf{0.1136}                                 & 0.9711                                            & 0.1455                                            \\ \hline
I\_IL   & 0.8951                                          & 0.4697                                          & 0.9403                                       & 0.2661                                      & 0.9052                                           & 0.4451                                          & \textbf{0.9449}                                   & \textbf{0.2358}                                   \\ \hline
I\_MB   & 0.9691                                          & 0.2691                                          & 0.9767                                       & 0.2109                                      & 0.9701                                           & 0.2592                                          & \textbf{0.9823}                                   & \textbf{0.1609}                                   \\ \hline
I\_BS   & 0.7892                                          & 0.4273                                          & 0.8643                                       & 0.2361                                      & 0.7772                                           & 0.4705                                          & \textbf{0.8758}                                   & \textbf{0.2250}                                   \\ \hline
O\_CL   & 0.9526                                          & 0.2411                                          & \textbf{0.9592}                              & \textbf{0.2050}                             & 0.9542                                           & 0.2359                                          & 0.9521                                            & 0.2459                                            \\ \hline
O\_RA   & 0.9423                                          & \textbf{0.0662}                                 & \textbf{0.9432}                              & 0.0809                                      & 0.9208                                           & 0.0936                                          & 0.9283                                            & 0.0944                                            \\ \hline
O\_SN   & \textbf{0.9279}                                 & \textbf{0.0581}                                 & 0.9124                                       & 0.0661                                      & 0.9017                                           & 0.0783                                          & 0.9055                                            & 0.0810                                            \\ \hline
O\_SU   & \textbf{0.9699}                                 & \textbf{0.0393}                                 & 0.9635                                       & 0.0467                                      & 0.9684                                           & 0.0413                                          & 0.9636                                            & 0.0469                                            \\ \hline
Average & 0.9299                                          & 0.2966                                          & 0.9390                                       & 0.2743                                      & 0.9307                                           & 0.2337                                          & \textbf{0.9412}                                   & \textbf{0.2070}                                   \\ \hline
\end{tabular}
}
\end{table}

\begin{table}[H]
 \caption{The comparison result of FM score on the LASIESTA dataset.}
  \centering
\scriptsize{
\begin{tabular}{ccccccccccc}
\hline
\begin{tabular}[c]{@{}c@{}}LASIESTA\\ DATASET\end{tabular} & \begin{tabular}[c]{@{}c@{}}Proposed\\ (SIE)\end{tabular} & \begin{tabular}[c]{@{}c@{}}Kim et al.\\ {[}1{]} (SIE)\end{tabular} & \begin{tabular}[c]{@{}c@{}}MSFgNet\\ {[}7{]} (SDE)\end{tabular} & \begin{tabular}[c]{@{}c@{}}Ortego\\ et al. {[}8{]}\end{tabular} & \begin{tabular}[c]{@{}c@{}}Fast-D\\ {[}9{]}\end{tabular} & \begin{tabular}[c]{@{}c@{}}SuBSENSE\\ {[}2{]}\end{tabular} & \begin{tabular}[c]{@{}c@{}}Berjón\\ et al. {[}10{]}\end{tabular} & \begin{tabular}[c]{@{}c@{}}3DCD\\ {[}30{]} (SIE)\end{tabular} & \begin{tabular}[c]{@{}c@{}}Haines\\ et al.\\ {[}11{]}\end{tabular} & \begin{tabular}[c]{@{}c@{}}FgSegNet\\ - v2 {[}16{]}\\ (SIE)\end{tabular} \\ \hline
I\_SI                                                      & 0.9301                                                   & \textbf{0.9617}                                                    & 0.9264                                                          & -                                                               & 0.9287                                                   & 0.9086                                                     & 0.8806                                                           & 0.8612                                                        & 0.8876                                                             & 0.6547                                                                   \\ \hline
I\_CA                                                      & \textbf{0.9589}                                          & 0.9145                                                             & 0.9213                                                          & -                                                               & 0.8924                                                   & 0.8702                                                     & 0.8444                                                           & 0.7156                                                        & 0.8938                                                             & 0.6411                                                                   \\ \hline
I\_OC                                                      & 0.9711                                                   & \textbf{0.9766}                                                    & 0.9163                                                          & -                                                               & 0.9194                                                   & 0.9249                                                     & 0.7807                                                           & 0.7512                                                        & 0.9223                                                             & 0.3995                                                                   \\ \hline
I\_IL                                                      & \textbf{0.9449}                                          & 0.8951                                                             & 0.8967                                                          & -                                                               & 0.5021                                                   & 0.4685                                                     & 0.6488                                                           & 0.8868                                                        & 0.8491                                                             & 0.4112                                                                   \\ \hline
I\_MB                                                      & \textbf{0.9823}                                          & 0.9691                                                             & 0.9143                                                          & -                                                               & 0.9430                                                   & 0.9173                                                     & 0.9374                                                           & 0.8274                                                        & 0.8440                                                             & 0.5291                                                                   \\ \hline
I\_BS                                                      & \textbf{0.8758}                                          & 0.7892                                                             & 0.7157                                                          & -                                                               & 0.6182                                                   & 0.6480                                                     & 0.6644                                                           & 0.8188                                                        & 0.6809                                                             & 0.3619                                                                   \\ \hline
O\_CL                                                      & 0.9521                                                   & \textbf{0.9526}                                                    & 0.8806                                                          & -                                                               & 0.9368                                                   & 0.9155                                                     & 0.9277                                                           & 0.8268                                                        & 0.8267                                                             & 0.5170                                                                   \\ \hline
O\_RA                                                      & 0.9283                                                   & \textbf{0.9424}                                                    & 0.8659                                                          & -                                                               & 0.9378                                                   & 0.8756                                                     & 0.8670                                                           & 0.8027                                                        & 0.8908                                                             & 0.4268                                                                   \\ \hline
O\_SN                                                      & 0.9055                                                   & \textbf{0.9280}                                                    & 0.8952                                                          & -                                                               & 0.8789                                                   & 0.7925                                                     & 0.7787                                                           & 0.6841                                                        & 0.1750                                                             & 0.1229                                                                   \\ \hline
O\_SU                                                      & 0.9636                                                   & \textbf{0.9699}                                                    & 0.7869                                                          & -                                                               & 0.8710                                                   & 0.7919                                                     & 0.7222                                                           & 0.8681                                                        & 0.8568                                                             & 0.1110                                                                   \\ \hline
Average                                                    & \textbf{0.9412}                                          & 0.9298                                                             & 0.8717                                                          & 0.8687                                                          & 0.8428                                                   & 0. 8113                                                    & 0.8051                                                           & 0.8043                                                        & 0.7826                                                             & 0.4175                                                                   \\ \hline
\end{tabular}
}
\end{table}

\begin{table}[H]
 \caption{Results by proposed algorithm on the LASIESTA dataset.}
  \centering
\small{
\begin{tabular}{cccccccc}
\hline
Category & FM     & PWC    & Recall & Precision & FPR    & FNR    & Sp      \\ \hline
I\_SI    & 0.9301 & 0.3263 & 0.9103 & 0.9620    & 0.0012 & 0.0897 & 0.9988  \\ \hline
I\_CA    & 0.9589 & 0.5089 & 0.9690 & 0.9500    & 0.0036 & 0.0310 & 0.9964  \\ \hline
I\_OC    & 0.9711 & 0.1455 & 0.9638 & 0.9787    & 0.0004 & 0.0362 & 0.9996  \\ \hline
I\_IL    & 0.9449 & 0.2358 & 0.9227 & 0.9684    & 0.0007 & 0.0773 & 0.9993  \\ \hline
I\_MB    & 0.9823 & 0.1609 & 0.9786 & 0.9862    & 0.0007 & 0.0214 & 0.9993  \\ \hline
I\_BS    & 0.8758 & 0.2250 & 0.9255 & 0.8389    & 0.0013 & 0.0745 & 0.9987  \\ \hline
O\_CL    & 0.9521 & 0.2459 & 0.9872 & 0.9216    & 0.0023 & 0.0128 & 0.9977  \\ \hline
O\_RA    & 0.9283 & 0.0944 & 0.918  & 0.9398    & 0.0002 & 0.0819 & 0.9998  \\ \hline
O\_SN    & 0.9055 & 0.0810 & 0.8824 & 0.9364    & 0.0004 & 0.1176 & 0.9996  \\ \hline
O\_SU    & 0.9636 & 0.0469 & 0.9722 & 0.9552    & 0.0003 & 0.0278 & 0.9997  \\ \hline
Average  & 0.9412 & 0.2070 & 0.9430 & 0.9437    & 0.0011 & 0.0570 & 0.9989 \\ \hline
\end{tabular}
}
\end{table}

\begin{figure}[H]
  \centering
  \includegraphics[scale=.45]{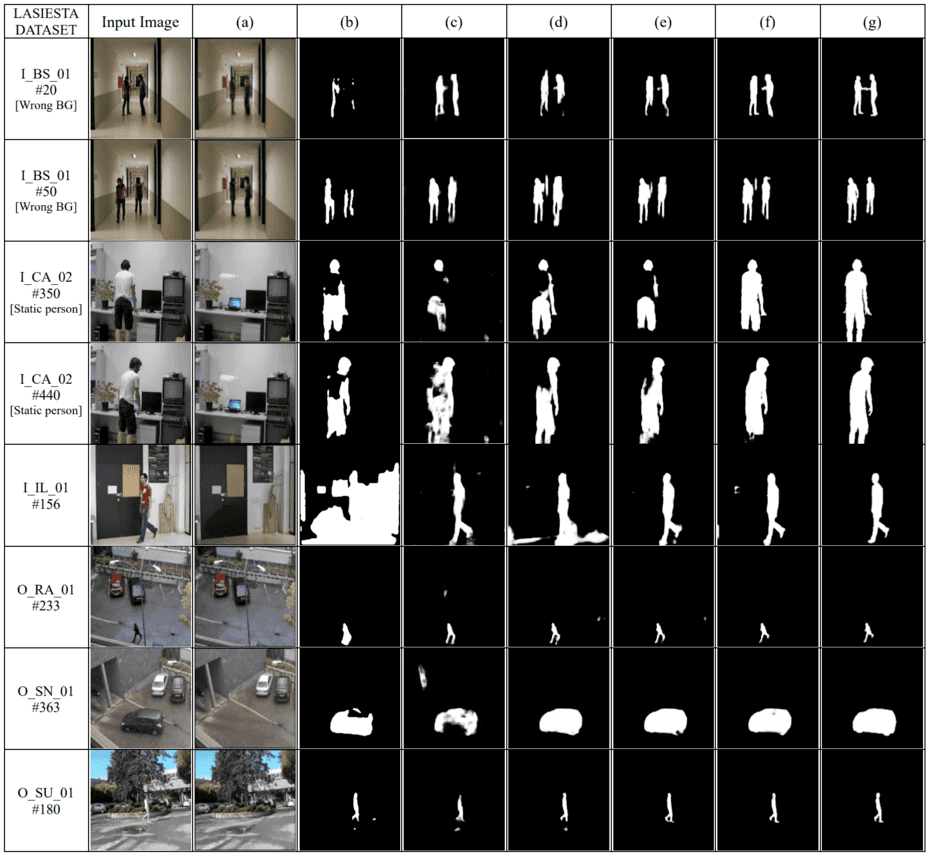}
  \caption{The comparison to other algorithms using LASIESTA dataset (a) background model image by SuBSENSE [2] (b) foreground map by SuBSENSE [2] (c) foreground map by 3DCD [30] (d) foreground map by our previous work [1] (e) foreground map by proposed algorithm using only background data augmentation (f) foreground map by proposed algorithm using both background and interval data augmentation (g) ground truth foreground map.}
  \label{fig:fig12}
\end{figure}

\subsubsection{Evaluation on Dataset Acquired Ourselves}

Evaluation was done using the dataset acquired ourselves and the tendency similar to the SBI and LASIESTA datasets was obtained. If there are no static human objects and wrong background model images, the improvement by the proposed data augmentation methods is unnoticeable. Even in a situation where the same background model images are used, the proposed method shows superior results compared to SuBSENSE [2]. Fig. 13 shows a comparison of detection performance in a rather simple environment in visual surveillance. Qualitative evaluation was done by SuBSENSE, 3DCD [30], Kim\&Ha [1], and proposed method. The methods excluding SuBSENSE were trained using 5,400 images of 27 scenes in the CDnet dataset like previous experiments. The proposed algorithm aims to improve the detection result for the case where SuBSENSE generates the wrong background model images and the foreground object that has been stopped for a long time. As shown in Fig. 13, in an environment where the background model image generated by the SuBSENSE is good and the foreground object is constantly moving, the proposed method shows slightly better performance than our previous work [1]. In the case of 3DCD [30], we can notice many cases of false detection which considers the background portion as the foreground. Also, the proposed method shows that the fragmentations effect is alleviated compared to existing studies.

In order to check the performance in the case of bootstrap and static human objects, additional datasets were acquired and evaluated. Fig. 14 shows comparison result of foreground map to other algorithms. We can notice clear improvement in this case compared to the case in Fig. 13. 3DCD [30] gives an incorrect result that consider background objects as foreground ones. Our previous study [1] also gives an incorrect result in this case. Background objects are erroneously detected as foreground ones when foreground objects are included in background model images. Also, person who is stationary for a long time is erroneously detected as background object. On the other hand, proposed algorithm gives good detection results in both cases of incorrect background model images are used input to a network and human objects stand still for a long time.

\begin{figure}[H]
  \centering
  \includegraphics[scale=.33]{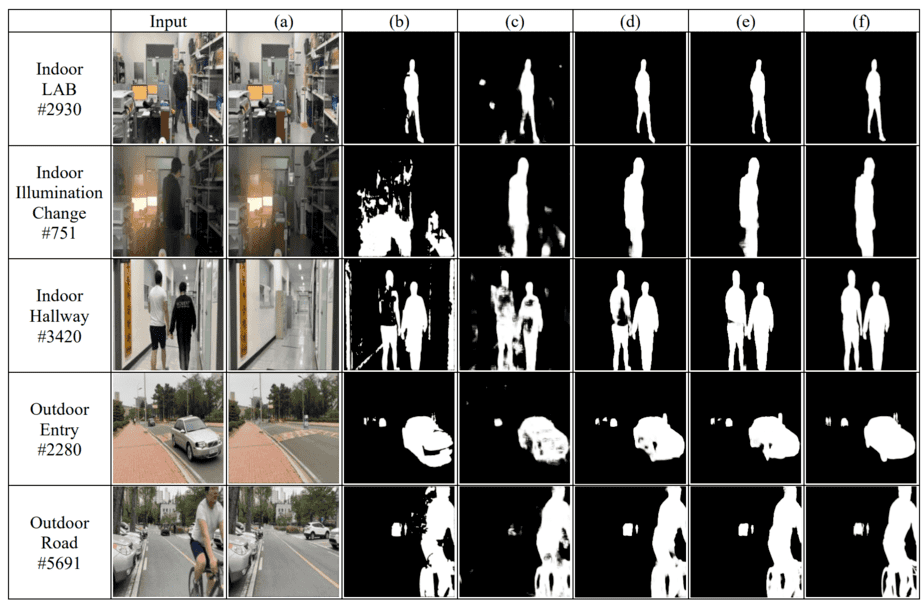}
  \caption{Comparison result of foreground maps using dataset acquired ourselves in easy cases (a) background model image by SuBSENSE [2] (b) SuBSENSE [2] (c) 3DCD [30] (d) Kim\&Ha [1], (e) proposed, (f) ground truth}
  \label{fig:fig13}
\end{figure}

\begin{figure}[H]
  \centering
  \includegraphics[scale=.33]{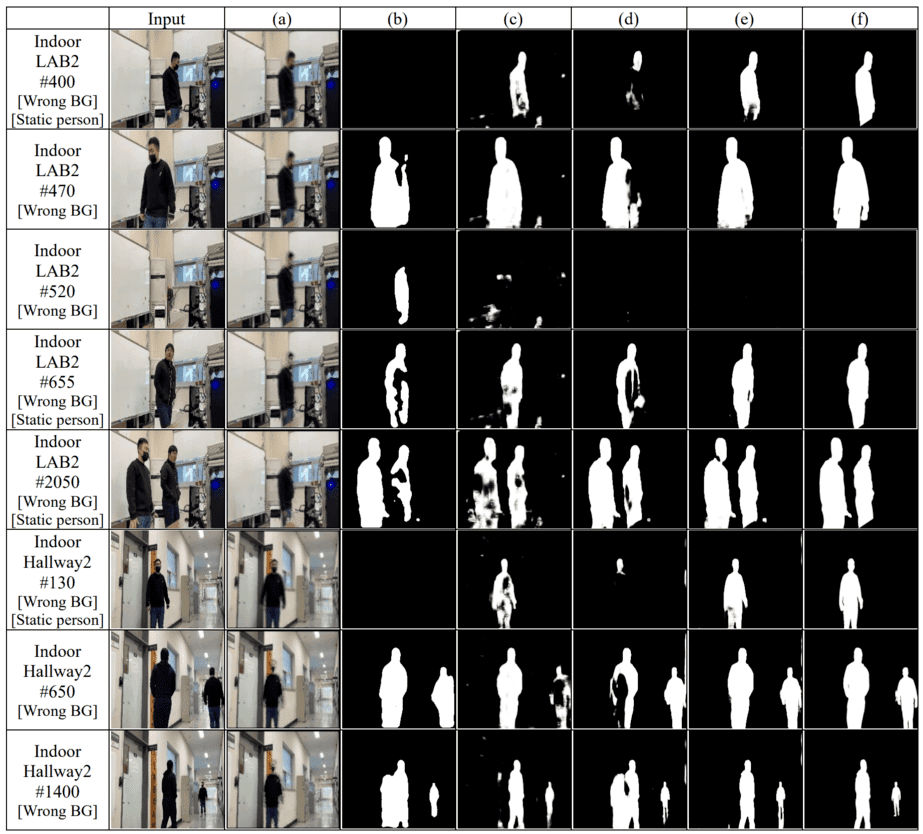}
  \caption{Comparison result of foreground maps using dataset acquired ourselves in difficult cases (a) background model image by SuBSENSE [2] (b) SuBSENSE [2] (c) 3DCD [30] (d) Kim\&Ha [1], (e) proposed, (f) ground truth}
  \label{fig:fig14}
\end{figure}

\subsection{Evaluation using training results with the SBI, LASIESTA dataset}

In this section, evaluation was done by dividing the SBI and LASIESTA datasets into the training and test set, while we tested using the SBI and LASIESTA datasets after training using the CDnet dataset in the previous section. In the case of the experiment using the SBI dataset, a total of 4 scenes of candela-m1.10, caviar2, CaVignal, and highway in the SBI dataset were used for test, and other scenes are used for training as in the experiments performed in [29] and [30]. In scenes of Foliage, Snellen, People\&Foliage and Toscana, moving leaves are classified as foreground objects contrary to the usual convention of visual surveillance that they should be classified as background objects. Therefore, we do not use those 4 scenes in the training. We think that this does not violate the criterion of test in SIE setup. Among the six scenes, except for the highway1 scene, all scenes are composed of only human. The number of people is about five times more than the number of vehicles. This bias in training data can cause problems. Therefore, training was done using only 100 or less images in the rest of the scenes except for the highway 1 scene to minimize the data bias problem. Table 8 shows the list of scenes and the number of images used for training for each scene. Table 9 shows a quantitative comparison to other deep learning-based methods using the SBI dataset in SIE setup. Compared algorithms used all data except for candela, caviar2, CaVignal, and highway2 in the SBI dataset for training. It can be seen that the proposed method shows higher performance than other algorithms even using with less data for training. The proposed algorithm gives a distinct improvement in candela and CaVignal scenes where there are incorrect background model images and people are stationary for a long time. The proposed algorithm gives the best result and shows 9.73\% improvement of FM score compared to our previous work [1].

Table 10 shows a quantitative comparison to other deep learning-based methods using the LASIESTA dataset in SIE setup. In the LASIESTA dataset, the evaluation was done by dividing each scene into training and test set. The LASIESTA dataset has 10 categories and each category has 2 scenes. Like the experiments performed in [29] and [30], the experiment was done in the SIE environment where training was done using the first scene of each category in the LASIESTA dataset, and evaluation was done using the second scenes. Ten scenes consist of 4,300 images, where 3,440 images are used for training and 860 images are used as validation. In the case of I-BS-1, I-CA-1, I-IL-1, I-MB-1, I-OC-1, I-SI-1 scenes, interval data augmentation can be used. If we use both background and interval data augmentation, we could have 12,450 images for training. Similar trends were obtained with the results using the SBI dataset. Both our previous work [1] and the proposed algorithm show better results than other deep learning methods. The proposed algorithm gives much better results compared to the SDE-based methods such as FgSegNet-S [31], FgSegNet-M [31], and FgSegNet-v2 [16]. Also, the proposed algorithm gives better results compared to the SIE-based methods such as ChangeDet [29] and 3DCD [30]. In the case of the LASIESTA dataset, the improvement through the proposed algorithm is not significant compared to our previous work [1] because background model images are good and there are almost no objects that have been stopped for a long time.

\begin{table}[H]
 \caption{Number of images used in training before and after data augmentation.}
  \centering
\begin{tabular}{|c|c|c|}
\hline
\begin{tabular}[c]{@{}c@{}}SBI DATASET\\ Scenes\end{tabular} & \begin{tabular}[c]{@{}c@{}}Before AUG\\ (Total/Train/Val)\end{tabular} & \begin{tabular}[c]{@{}c@{}}After AUG\\ (Total/Train/Val)\end{tabular} \\ \hline
Board                                                        & 100/80 /20                                                             & 400/320/80                                                            \\ \hline
CAVIAR1                                                      & 100/80/20                                                              & 400/320/80                                                            \\ \hline
Hall\&Monitor                                                & 100/80/20                                                              & 400/320/80                                                            \\ \hline
Highway1                                                     & 439/351/88                                                             & 878/702/176                                                           \\ \hline
HumanBody2                                                   & 100/80/20                                                              & 400/320/80                                                            \\ \hline
IBMtest2                                                     & 90/72/18                                                               & 360/288/72                                                            \\ \hline
Total                                                        & 929/743/186                                                            & 2838/2270/568                                                         \\ \hline
\end{tabular}
\end{table}

\begin{table}[H]
 \caption{Comparison of FM score using SBI dataset under SIE environments.}
  \centering
\begin{tabular}{cccccccc}
\hline
\begin{tabular}[c]{@{}c@{}}SBI\\ DATASET\end{tabular} & \begin{tabular}[c]{@{}c@{}}FgSegNet-S\\ {[}31{]}\end{tabular} & \begin{tabular}[c]{@{}c@{}}FgSegNet-M\\ {[}31{]}\end{tabular} & \begin{tabular}[c]{@{}c@{}}FgSegNet- v2\\ {[}16{]}\end{tabular} & \begin{tabular}[c]{@{}c@{}}ChangeDet\\ {[}29{]}\end{tabular} & \begin{tabular}[c]{@{}c@{}}3DCD\\ {[}30{]}\end{tabular} & \begin{tabular}[c]{@{}c@{}}Kim\&Ha\\ {[}1{]}\end{tabular} & Proposed      \\ \hline
Candela                                               & 0.23                                                          & 0.15                                                          & 0.27                                                            & 0.61                                                         & 0.67                                                    & 0.64                                                      & \textbf{0.70} \\ \hline
CAVIAR2                                               & 0.11                                                          & 0.14                                                          & 0.10                                                            & 0.56                                                         & 0.62                                                    & \textbf{0.94}                                             & 0.92          \\ \hline
CaVignal                                              & 0.68                                                          & 0.72                                                          & 0.63                                                            & 0.48                                                         & 0.53                                                    & 0.46                                                      & \textbf{0.73} \\ \hline
Highway2                                              & 0.24                                                          & 0.21                                                          & 0.58                                                            & 0.64                                                         & 0.59                                                    & \textbf{0.98}                                             & 0.96          \\ \hline
Avg                                                   & 0.31                                                          & 0.30                                                          & 0.40                                                            & 0.57                                                         & 0.60                                                    & 0.75                                                      & \textbf{0.83} \\ \hline
\end{tabular}
\end{table}

\begin{table}
 \caption{Comparison of FM score using LASIESTA dataset under SIE environments.}
  \centering
\scriptsize{
\begin{tabular}{cccccccc}
\hline
\begin{tabular}[c]{@{}c@{}}LASIESTA\\ DATASET\end{tabular} & \begin{tabular}[c]{@{}c@{}}FgSegNet-S\\ {[}31{]}\end{tabular} & \begin{tabular}[c]{@{}c@{}}FgSegNet-M\\ {[}31{]}\end{tabular} & \begin{tabular}[c]{@{}c@{}}FgSegNet-v2\\ {[}16{]}\end{tabular} & \begin{tabular}[c]{@{}c@{}}ChangeDet\\ {[}29{]}\end{tabular} & \begin{tabular}[c]{@{}c@{}}3DCD\\ {[}30{]}\end{tabular} & \begin{tabular}[c]{@{}c@{}}Kim\&Ha\\ {[}1{]}\end{tabular} & Proposed      \\ \hline
I\_SI\#2                                                   & 0.20                                                          & 0.56                                                          & 0.53                                                           & 0.83                                                         & 0.86                                                    & 0.96                                                      & \textbf{0.98} \\ \hline
I\_CA\#2                                                   & 0.60                                                          & 0.55                                                          & 0.58                                                           & 0.55                                                         & 0.49                                                    & 0.89                                                      & \textbf{0.92} \\ \hline
I\_OC\#2                                                   & 0.53                                                          & 0.65                                                          & 0.25                                                           & 0.89                                                         & 0.93                                                    & \textbf{0.97}                                             & 0.96          \\ \hline
I\_IL\#2                                                   & 0.25                                                          & 0.42                                                          & 0.41                                                           & 0.87                                                         & 0.85                                                    & 0.95                                                      & \textbf{0.96} \\ \hline
I\_MB\#2                                                   & 0.60                                                          & 0.56                                                          & 0.63                                                           & 0.77                                                         & 0.79                                                    & 0.92                                                      & \textbf{0.97} \\ \hline
I\_BS\#2                                                   & 0.28                                                          & 0.19                                                          & 0.25                                                           & \textbf{0.82}                                                & 0.87                                                    & 0.68                                                      & 0.74          \\ \hline
O\_CL\#2                                                   & 0.19                                                          & 0.28                                                          & 0.54                                                           & 0.90                                                         & 0.87                                                    & \textbf{0.97}                                             & 0.97          \\ \hline
O\_RA\#2                                                   & 0.16                                                          & 0.18                                                          & 0.54                                                           & 0.85                                                         & 0.87                                                    & 0.98                                                      & \textbf{0.98} \\ \hline
O\_SN\#2                                                   & 0.05                                                          & 0.01                                                          & 0.05                                                           & 0.51                                                         & 0.49                                                    & \textbf{0.83}                                             & 0.81          \\ \hline
O\_SU\#2                                                   & 0.18                                                          & 0.33                                                          & 0.29                                                           & 0.81                                                         & 0.83                                                    & 0.92                                                      & \textbf{0.94} \\ \hline
Avg                                                        & 0.30                                                          & 0.37                                                          & 0.41                                                           & 0.79                                                         & 0.79                                                    & 0.91                                                      & \textbf{0.92} \\ \hline
\end{tabular}
}
\end{table}

\subsection{Layer visualization}

In order to find out the effect of the proposed data augmentation methods, the output of each layer in the U-NET structure was visualized and analyzed. In this paper, we propose data augmentation methods for improving the performance of our previous work [1], so both the proposed method and the previous study use the same U-NET model. Fig. 15 shows fives layers which were selected for analysis. They correspond to layers close to the input and output, and we selected them because they are suitable for comparing the effect of using data augmentation. Since the role of each layer is different, only the trends by the layer are analyzed through layer visualization according to the use of the augmentation method. Since each layer is composed of multiple channels such as 64, 128, 256 channels, it is difficult to visualize the entire channels at once. Therefore, specific channels in each layer was manually selected and their activation maps were converted into gray images for visualization. Fig. 16 shows various activation maps for selected channels in a layer. We can notice that general feature extraction is done in layers close to inputs and there is no significant difference between the proposed algorithm and our previous work [1]. However, we can notice clear effects by the proposed data augmentation methods at layers close to outputs. Incorrectly classified foreground objects which are included in the background model images are correctly discriminated by the proposed data augmentation methods. Fig. 17 shows activation maps of various channels in a channel close to the output using CaVignal \#200 image. Although our previous work [1] and the proposed algorithm use the same network, we can notice distinct difference in those activation maps by the proposed data augmentation methods.

Since there are layers that respond to the foreground object and layers that respond to the background object depending on the layer, it was judged that what value is displayed in the foreground object is not an important factor. From the investigation of the various activation maps in Fig. 16 and Fig. 17, we can conclude that there are some specific channels which activates for background and foreground objects. For better understanding, we analyzed the difference between activation maps by people included in background model images and original images. When data augmentation is not used, the model could not discriminate foreground objects which are falsely included in background model images. We can notice that the model can cope with incorrect foreground objects which are included in background model images if we use proposed data augmentation methods. We can indirectly confirm that the proposed data augmentation methods improve foreground objects detection ability because both the proposed and our previous work uses the same model. Through this, we can conclude that improvements in visual surveillance are possible by both data augmentation and modification of the structure of the deep learning model.

\begin{figure}[H]
  \centering
  \includegraphics[scale=.11]{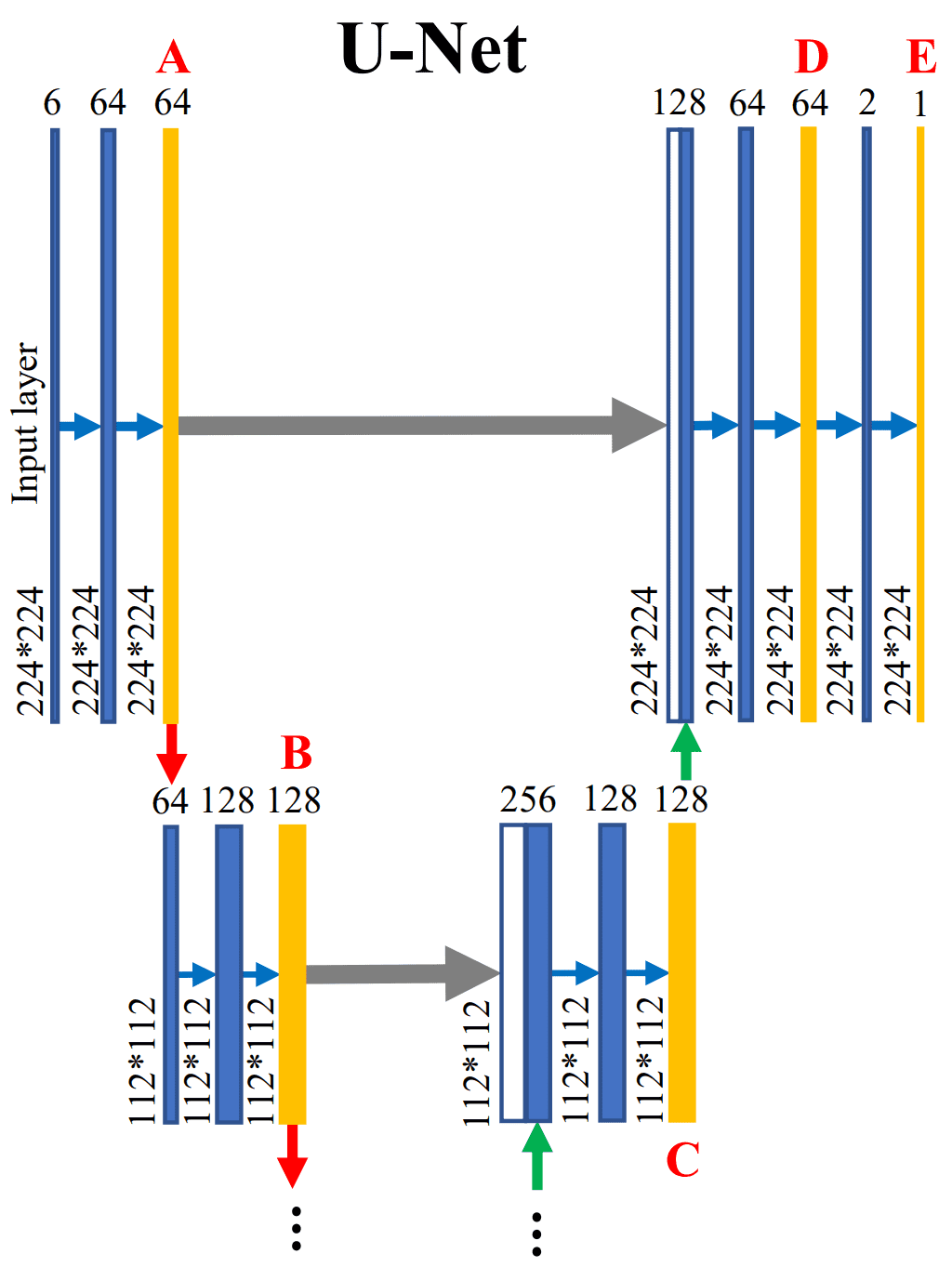}
  \caption{Layers selected for layer visualization and they are display in yellow.}
  \label{fig:fig15}
\end{figure}

\begin{figure}[H]
  \centering
  \includegraphics[scale=.47]{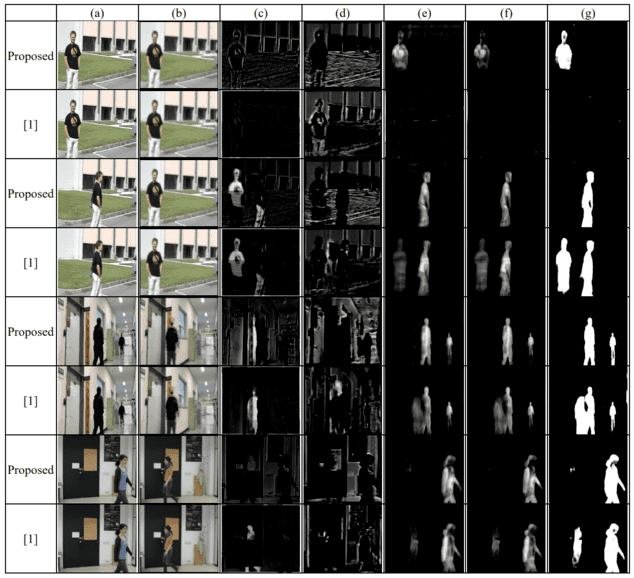}
  \caption{Comparison of activation map at various layer and channel (a) input image (b) background model image by SuBSENSE (c) layer A channel 38 (d) layer B channel 125 (e) layer C channel 111 (f) layer D channel 59 (g) layer E channel 1.}
  \label{fig:fig16}
\end{figure}

\begin{figure}[H]
  \centering
  \includegraphics[scale=.47]{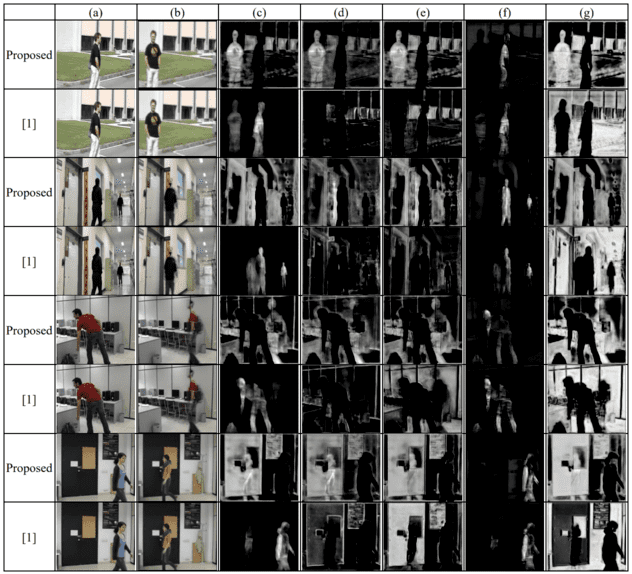}
  \caption{Comparison of activation map at layer D (a) input image (b) background model image by SuBSENSE (c) channel 7 (d) channel 13 (e) channel 34 (f) channel 55 (g) channel 57.}
  \label{fig:fig17}
\end{figure}

\section{Conclusion}
\label{sec:others}

In this paper, we proposed data augmentation methods suitable for the network that uses spatio-temporal inputs of a number of past, current, and background model images. In deep learning, data augmentation techniques are widely used in image classification and object detection, and most of them target one image. We proposed data augmentation methods in the spatio-temporal domain by adjusting the frame interval and modifying the background model images. Through the proposed method, it is possible to improve the detection performance of the ghost objects problem that can occur when the BGS algorithm creates an incorrect background model images in the bootstrap environment and the human foreground objects that are stopped for a long time. Through quantitative and qualitative comparisons using the SBI, LASIESTA, and our own dataset, it showed superior performance compared to the latest deep learning algorithms and traditional BGS algorithms. Unlike deep learning-based methods designed for SDE environments, the proposed method shows excellent performance in environments different from training without requiring additional labels and re-training. Also, the proposed algorithm in this paper uses a simple U-NET structure. Nevertheless, it can be seen that it shows superior results compared to the latest deep learning methods and traditional BGS algorithms. Therefore, research not only on deep learning models but also on the configuration of inputs and their data augmentation is important in visual surveillance. In the future research, we plan to improve the accuracy of the model by changing the U-NET and replace SuBSENSE with a CNN which estimates a background model image.

% \begin{itemize}
% \item Lorem ipsum dolor sit amet
% \item consectetur adipiscing elit. 
% \item Aliquam dignissim blandit est, in dictum tortor gravida eget. In ac rutrum magna.
% \end{itemize}

\section*{Acknowledgement}
This work was supported by the National Research Foundation of Korea (NRF) grant funded by the Korea government. (MSIT) (2020R1A2C1013335).

\bibliographystyle{unsrt}  


\end{document}